\documentclass[sigconf]{acmart}
\usepackage{booktabs} % To thicken table lines

\usepackage[english]{babel}
\usepackage{moresize}
\usepackage{amsmath}
\usepackage{algorithmic}
\usepackage{balance}
\usepackage{comment}
\usepackage{paralist}
\usepackage{bm}
\usepackage{pgfplots}
\usetikzlibrary{pgfplots.dateplot}

\usepackage{flushend}
\usepackage[english]{babel}
\usepackage[latin1]{inputenc}
\usepackage{mathrsfs}
\usepackage{graphicx}

\usepackage{amssymb}
\usepackage{amsfonts}
\usepackage{url}
\usepackage{longtable}
\usepackage{rotating}
\usepackage{multirow}
\usepackage{subfigure}
\usepackage{enumitem}
\usepackage[linesnumbered,algoruled,boxed,lined]{algorithm2e}
\usepackage{adjustbox}
\usepackage{hyperref}
\usepackage{pgfplots}
\usetikzlibrary{pgfplots.dateplot}
\usepackage{filecontents}
\usepackage{diagbox}

\definecolor{tblue}{RGB}{31,119,180}
\definecolor{torange}{RGB}{255,127,14}
\definecolor{tgreen}{RGB}{44,160,44}
\definecolor{tred}{RGB}{214,39,40}
\definecolor{tpurple}{RGB}{148,103,189}

\newcommand{\hide}[1]{} %hide

\newcommand{\eg}{\textit{e}.\textit{g}.}

\def\model{OpenCity}

\setcopyright{none}
\begin{document}
\fancyhead{}

\title{OpenCity: Open Spatio-Temporal Foundation Models \\ for Traffic Prediction}

\author{Zhonghang Li$^{1,2}$, Long Xia$^3$, Lei Shi$^3$, Yong Xu$^2$, Dawei Yin$^3$ and Chao Huang$^{1*}$}
\thanks{$*$ Chao Huang is the Corresponding Author.}
\affiliation{$^1$The University of Hong Kong, $^2$South China University of Technology, $^3$Baidu Inc.\\}
% \affiliation{\textbf{Project Page}: \href{https://urban-gpt.github.io/}{https://OpenCity-ST.github.io/}, \textbf{Github}: \href{https://github.com/HKUDS/OpenCity}{https://github.com/HKUDS/OpenCity}}


\begin{abstract}
Accurate traffic forecasting is crucial for effective urban planning and transportation management, enabling efficient resource allocation and enhanced travel experiences. However, existing models often face limitations in generalization, struggling with zero-shot prediction on unseen regions and cities, as well as diminished long-term accuracy. This is primarily due to the inherent challenges in handling the spatial and temporal heterogeneity of traffic data, coupled with the significant distribution shift across time and space. In this work, we aim to unlock new possibilities for building versatile, resilient and adaptive spatio-temporal foundation models for traffic prediction. To achieve this goal, we introduce a novel foundation model, named \model, that can effectively capture and normalize the underlying spatio-temporal patterns from diverse data characteristics, facilitating zero-shot generalization across diverse urban environments. \model\ integrates the Transformer architecture with graph neural networks to model the complex spatio-temporal dependencies in traffic data. By pre-training \model\ on large-scale, heterogeneous traffic datasets, we enable the model to learn rich, generalizable representations that can be seamlessly applied to a wide range of traffic forecasting scenarios. Experimental results demonstrate that \model\ exhibits exceptional zero-shot predictive performance. Moreover, \model\ showcases promising scaling laws, suggesting the potential for developing a truly one-for-all traffic prediction solution that can adapt to new urban contexts with minimal overhead. 
% Source codes are available.
We made our proposed \model\ model open-source and it is available at the following link: {\color{blue}\url{https://github.com/HKUDS/OpenCity}}.
\end{abstract}



% \keywords{datasets, neural networks, gaze detection, text tagging}

\maketitle

\section{Introduction}
\label{sec:intro}

Transportation is an essential component of urban activities, serving as the fundamental infrastructure that supports the efficient movement of people and goods within cities~\cite{tedjopurnomo2020survey,yin2021deep}. Precise traffic forecasting allows for proactive traffic management, enabling transportation planners to anticipate and mitigate potential congestion, delays, and disruptions~\cite{jin2023spatio,lei2022modeling}. This, in turn, enhances the overall efficiency and sustainability of urban transportation networks, fostering the real-world intelligent transportation systems.

The success of deep learning has empowered spatio-temporal models for traffic forecasting. These models leverage deep neural networks to learn effective representations that can better capture the spatial and temporal dependencies inherent in urban traffic data. However, despite these advancements, current traffic prediction models often face significant limitations in terms of generalization. \\\vspace{-0.12in}

\noindent \textbf{Cross-Regional Model Spatial Generalization}. Firstly, A key limitation of current traffic prediction models is their struggle with spatial generalization. These models often fail to perform well when applied to unseen regions or cities, as traffic patterns and dynamics can vary considerably across geographical locations due to factors like infrastructure, demographics, and urban planning~\cite{bai2020adaptive}. Existing models typically learn from data limited to specific regions, making them unable to effectively generalize their knowledge to capture the unique characteristics of unfamiliar traffic environments. 

This issue is critical, as it is impractical to deploy comprehensive sensor networks across entire urban areas to collect traffic data~\cite{jin2022selective,lu2022spatio}. A more viable approach is to build models that can generalize well to unseen regions using only partial data. Additionally, developing spatio-temporal models that are applicable across different cities would significantly reduce deployment and maintenance costs~\cite{jin2022selective,yao2019learning}. However, when these models are applied in new locations, they often experience a significant drop in predictive performance, hindering their wider applicability. This spatial generalization challenge is crucial to address in order to create traffic forecasting solutions that can be seamlessly deployed across diverse urban settings without extensive retraining or fine-tuning.
\\\vspace{-0.12in}

\noindent \textbf{Temporal Generalization for Long-term Forecasting}. Current traffic prediction models excel at short-term forecasting, such as anticipating conditions for the next hour~\cite{cirstea2022towards,jiang2023pdformer}. However, their ability to generalize to longer time frames, like days or weeks ahead, is notably limited. This limitation is largely due to the models' poor generalization ability in effectively handling the evolving temporal distribution shifts that occur over longer time horizons in practical urban scenarios. As the forecasting timeframe increases, these models struggle to capture and account for the dynamic changes in traffic patterns that influence long-term traffic conditions.

This limitation presents a considerable obstacle for city planners and transportation agencies striving to devise effective long-term strategies. Precise long-term traffic forecasts are crucial for a range of applications, including proactive infrastructure planning, public transit scheduling, and event logistics coordination. Despite their importance, existing models often fall short when it comes to providing reliable predictions beyond the immediate future.

In recent years, the remarkable capabilities of large foundation models have captured significant attention across various domains. These models, trained on vast and diverse datasets, have demonstrated exceptional performance in natural language processing (NLP)~\cite{brown2020language,ouyang2022training} and computer vision (CV)~\cite{he2022masked,kirillov2023segment} tasks, showcasing their remarkable abilities to comprehend and generalize from complex data. While the success of foundation models has been well-documented in the NLP and CV realms, the development of a foundation model tailored specifically for urban traffic data remains largely unexplored. In this study, we aim to bridge this gap by creating a novel foundation model that is specifically designed to capture the complexities of urban traffic. Our primary focus is on enhancing the model's ability to generalize across both spatial and temporal dimensions, enabling it to perform robust cross-regional traffic forecasting and long-term temporal predictions. \\\vspace{-0.12in}

\begin{figure}[t]
    \centering
    % \vspace{-0.1in}
    \subfigure{
        % \vspace{-0.08in}
        \begin{minipage}[t]{0.5\linewidth}
            \centering
            \includegraphics[width=1.6in]{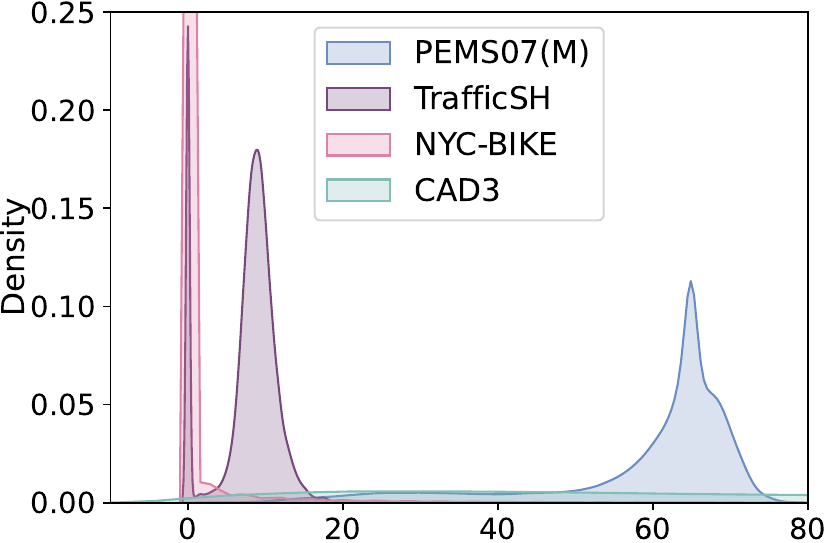}
        \end{minipage}%
        \begin{minipage}[t]{0.50\linewidth}
            \centering
            \includegraphics[width=1.60in]{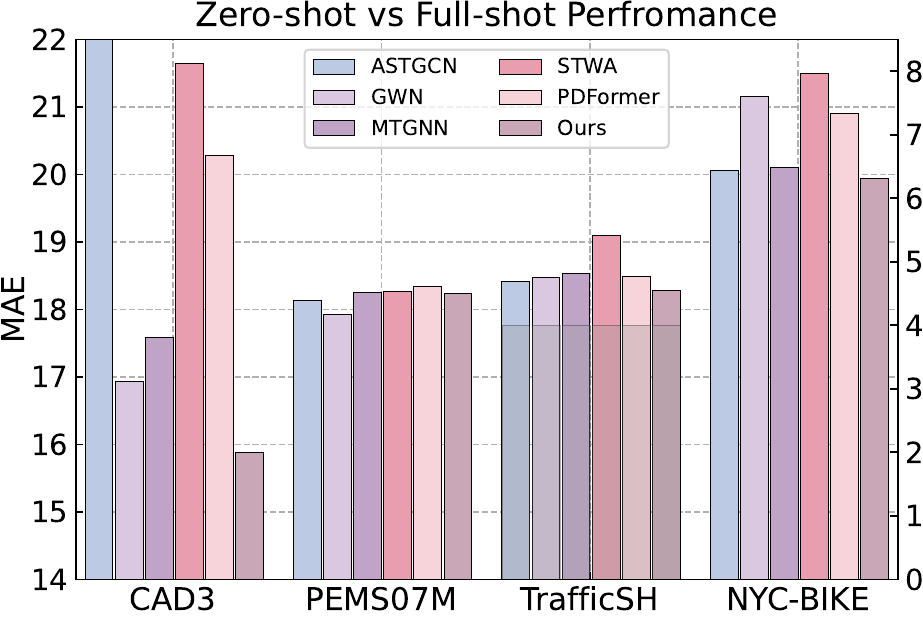}
        \end{minipage}%
    }%
    \centering
    \vspace{-0.2in}
    % \caption{Motivations of \model. The left figure illustrates the varying distributions across different traffic datasets, while the right figure presents the zero-shot (\model) vs. full-shot (baselines) performance on these datasets. Despite challenges, \model\ performs comparably to baselines.}
    \caption{The left figure showcases varying data distributions across traffic datasets, highlighting the need for a model that can handle such disparities. The right figure compares the zero-shot \model\ performance to full-shot baselines, showing \model\ performs comparably despite the distribution shift challenges from spatio-temporal heterogeneity.}
    \vspace{-0.17in}
    \label{fig:intro}
\end{figure}

\noindent \textbf{Contribution}. In this work, we introduce the One-for-All Spatio-Temporal Model-\model, a novel and versatile framework for developing foundational urban traffic models with powerful zero-shot learning capabilities. The primary objective of \model\ is to address the limitations of existing traffic forecasting approaches, which often struggle to generalize beyond their initial training domains. By providing a unified modeling architecture, \model\ aims to enable seamless adaptation across a wide range of traffic conditions and contexts. Achieving this goal of versatility and adaptability, however, presents several critical challenges:
\begin{itemize}[leftmargin=*]

\item \textbf{Learning Effective Universal Representations}. The ability to learn powerful, generalizable representations of traffic dynamics is crucial for developing a versatile traffic model. This challenge entails capturing the essential patterns and relationships underlying urban mobility, while minimizing the reliance on context-specific features that may impede the model's adaptability to new environments. By learning universal representations, the model can acquire a robust and transferable understanding of traffic patterns, enabling it to be effectively applied to a diverse range of scenarios, even without domain-specific training data. \\\vspace{-0.12in}

\item \textbf{Confronting Spatio-Temporal Heterogeneity}. Urban traffic patterns are characterized by their inherent diversity, showing notable variations in distribution across various spatial areas and timeframes. Addressing this heterogeneity is essential for ensuring that a unified spatio-temporal traffic model remains versatile and adaptable. For a model to seamlessly adjust to the multitude of traffic scenarios it may encounter, it must be crafted to adeptly manage these shifts in distribution.

\end{itemize}

To address the key challenges, the proposed model, \model, integrates the Transformer architecture and graph neural networks. This hybrid approach enables the model to effectively capture the complex spatio-temporal dependencies inherent in traffic data. At the core of \model\ is the seamless cooperation between three key components: a zero-shot spatio-temporal embedding layer, a spatio-temporal context encoder, and spatio-temporal dependency modeling components. This architecture allows the model to learn universal representations that can generalize across diverse spatial and temporal contexts. Furthermore, the model is strengthened through pre-training on large-scale, heterogeneous traffic datasets, equipping it with rich and transferable representations. In summary, this work makes the following major contributions:

\begin{itemize}[leftmargin=*]

\item \textbf{Versatile Spatio-Temporal Modeling}. \model\ is designed to effectively handle the inherent diversity and variations in urban traffic patterns across different spatial areas and timeframes. By confronting this spatio-temporal heterogeneity, \model\ overcomes the reliance on context-specific features that often hinders the generalization capabilities of current spatio-temporal models. \\\vspace{-0.12in}

\item \textbf{Impressive Zero-Shot Forecasting Capabilities}. The \model\ demonstrates superior performance compared to \textbf{full-shot} models trained exclusively on target domains.  This remarkable \textbf{zero-shot} capability highlights the model's ability to learn generalizable representations, enabling seamless application to new traffic environments without extensive retraining or fine-tuning. \\\vspace{-0.12in}

\item \textbf{Fast Context-Adaptive Ability}. The remarkable adaptability of our \model\ suggests its broad applicability across a diverse range of spatio-temporal prediction tasks. By requiring only efficient and expedited fine-tuning for context adaptation, the model can be seamlessly deployed in a wide variety of scenarios. \\\vspace{-0.12in}

\item \textbf{Scaling Laws Investigation}. \model\ showcases promising scaling laws, suggesting the potential for effectively scaling and adapting to new, previously unseen scenarios with minimal additional training or fine-tuning required across diverse contexts.

\end{itemize}

% \vspace{-0.10in}
\section{Preliminaries}
\label{sec:model}

\textbf{Spatio and Temporal Unit Generation}.
Traffic data is inherently spatio-temporal in nature, reflecting the dynamic patterns and spatial distribution of transportation networks. It is typically represented as a two-dimensional matrix $\textbf{X}\in\mathbb{R}^{R\times T}$, where each element corresponds to a specific traffic metric (\eg, flow, speed, or demand), for a given region $r$ and time interval $t$. There are two prevalent forms of spatial region unit used to model traffic data: \vspace{-0.05in}

\begin{itemize}[leftmargin=*]
    \item i) \textbf{Sensor-based Traffic Network}: The traffic data is collected through an irregular sensor network across major roads. Each sensor represents a region, and the region relationships are modeled using a graph $G = (V, E, A)$, where $V$ are regions, $E$ are edges, and $A \in \mathbb{R}^{R \times R}$ is a weighted adjacency matrix capturing spatial dependencies, based on geographical distance. \\\vspace{-0.12in}
    
    \item ii) \textbf{Grid-based Traffic Network}: the urban area is partitioned into a regular grid of uniform square blocks, often with dimensions of 1km $\times$ 1km. This grid-like spatial representation allows the transportation network to be modeled using a graph structure, where each grid block is represented as a node, and the connections between neighboring regions are captured as edges. \vspace{-0.1in}
\end{itemize}

\noindent \textbf{Traffic Flow Forecasting.}
The task of traffic forecasting is to use data from the past $H$ time steps ($\textbf{X}_{t_{I-H+1}: t_I}$) to predict traffic data $\textbf{Y}_{t_{I+1}: t_{I+F}}$ for the future $F$ time steps. This involves exploring the complex spatial and temporal patterns inherent in traffic data to uncover the underlying dynamics that drive changes in traffic flow.
\begin{align}
    \label{eq:Preliminarie1}
    \textbf{Y}_{t_{I+1}: t_{I+F}} = f(\textbf{X}_{t_{I-H+1}: t_I})
\end{align}
where $f(\cdot)$ represents the spatio-temporal prediction function. \\\vspace{-0.1in}

% In this work, we focus on long-term traffic prediction, such as forecasting traffic conditions for the following day rather than just the next hour. This task not only incorporates the results of short-term predictions but also provides more comprehensive insights into future traffic trends, better fulfilling the needs of government management and public travel planning decisions.

\noindent \textbf{Spatio-Temporal Generalization}. In this work, we propose a foundation model to empower traffic prediction with strong generalization capabilities across both spatial and temporal dimensions. \\\vspace{-0.12in}

\noindent (i) \textbf{Temporal Generalization for Long-term Forecasting}. Most existing traffic prediction models are limited to forecasting only short-term (\eg, next hour) traffic variations. These models often struggle to generalize well to long-term traffic forecasting due to their poor temporal generalization ability. However, in practical applications, long-term traffic forecasting (\eg, days or weeks into the future) is critically important and highly beneficial for various transportation planning and management tasks. \\\vspace{-0.12in}

\noindent (ii) \textbf{Spatial Generalization for Zero-Shot Forecasting}.
In real-world intelligent transportation systems, data heterogeneity and scarcity pose significant challenges. For instance, traffic flow patterns can vary substantially across different geographical regions within a city, and also differ considerably between cities. This presents a major obstacle for current traffic flow prediction models, as they often struggle to generalize their performance to new locations where little or no historical data is available.

% Zero-shot traffic prediction is a crucial task for assessing the performance of a foundation model. This involves evaluating the model on a test set that has no overlap in region or traffic metrics with the pre-training dataset. For example, using traffic flow data from City A as the pre-training dataset and traffic flow data from City B as the test dataset. Alternatively, using taxi demand data from City A as the pre-training dataset and bicycle demand data from the same city as the test dataset. Zero-shot prediction process can be formalized as follows:

Given the discussion above, we can formally define the traffic flow prediction task with strong spatio- temporal generalization:
\begin{align}
    \label{eq:Preliminarie2}
    \tilde{\textbf{Y}}_{t_{I+1}: t_{I+F}} = \Bar{f}(\tilde{\textbf{X}}_{t_{I-H+1}: t_I})
\end{align}
Different from the predictive function ($f(\cdot)$) in existing spatio-temporal models, the proposed $\Bar{f}(\cdot)$ is a generalized spatio-temporal foundation model. This model can be directly utilized to make predictions on downstream unseen traffic data $\tilde{\textbf{X}}$, which has no overlap with the training data $\textbf{X}$. $\Bar{f}(\cdot)$ is pre-trained on a diverse set of traffic data sources. It captures the underlying patterns and relationships that govern traffic dynamics across different spatial and temporal contexts. This enables zero-shot traffic flow forecasting without requiring any additional fine-tuning or adaptation steps.

% The function $\Bar{f}$ keeps its parameters frozen, and $\tilde{\textbf{X}}$ shares no spatial or statistical features with $\textbf{X}$ under this scenario.

\section{Methodology}
\label{sec:solution}

\begin{figure*}[t]
    \centering
    \includegraphics[width=2.1\columnwidth]{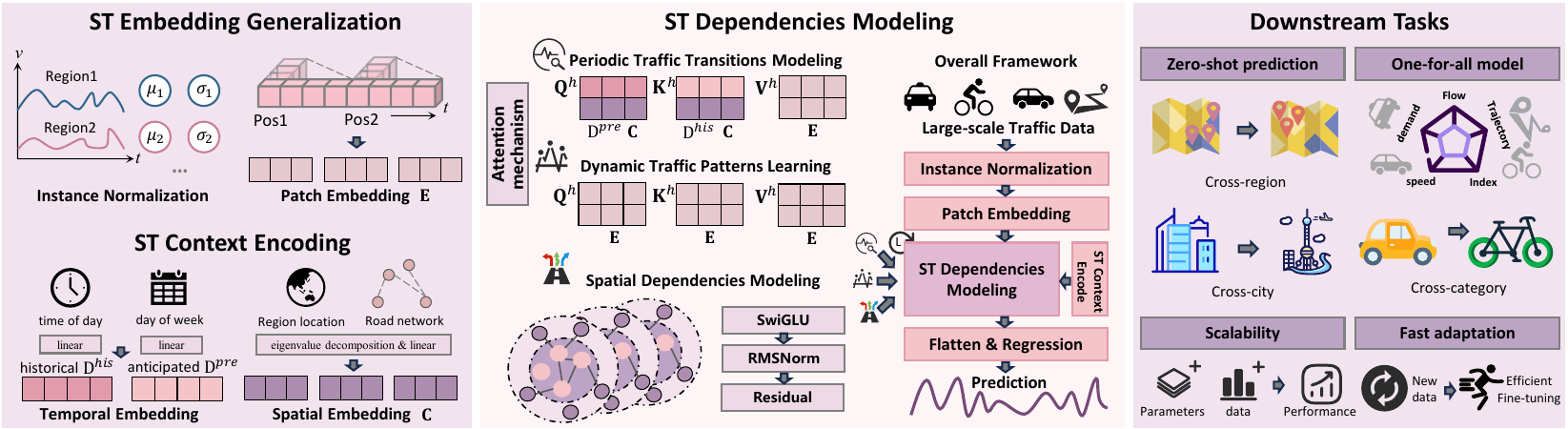}
    \vspace{-0.2in}
    \caption{The overall architecture of Spatio-Temporal Foundation Model \model.}
    \vspace{-0.1in}
    \label{fig:fra01}
\end{figure*}

\vspace{-0.05in}
\subsection{Spatio-Temporal Embedding for Distribution Shift Generalization}
% We initiate the processing of spatio-temporal data by employing a spatio-temporal data embedding layer, which consists of two primary steps: instance normalization and patch embedding.
The design of our spatio-temporal embedding layer is driven by the need to address the distribution shift across both spatial and temporal dimensions in traffic prediction tasks. In real-world intelligent transportation systems, traffic flow patterns can vary considerably across different geographical regions and time periods. This inherent data heterogeneity presents a major obstacle for traffic prediction models, as they need to generalize their performance to new environments or scenarios where the underlying data distribution may be drastically different from the training data.

\vspace{-0.05in}
\subsubsection{\bf In-Context Normalization for Zero-Shot Generalization} 
Existing approaches often leverage statistical features of the training data, such as mean and standard deviation, for data normalization. However, these summary statistics may be inadequate or non-transferable when the test data exhibits significant data heterogeneity and has no spatial overlap with the training data distribution. To address this challenge and accommodate zero-shot traffic prediction tasks, we employ instance normalization $\textbf{IN}(\cdot)$ to process the data. This approach utilizes the mean and standard deviation of the individual input instances $\textbf{X}_r\in\mathbb{R}^{T}$ for each region, rather than relying on the global training set statistics. The formalization of this process is as follows:
\begin{align}
    \label{eq:IN}
    \bar{\textbf{X}}_{r,t} = \textbf{IN}(\textbf{X}_{r,t}) = \frac{\textbf{X}_{r,t}-\mu_r}{\sigma_r}
\end{align}
\noindent Where $\mu(\textbf{X}_r)$ and $\sigma(\textbf{X}_r)$ are the mean and standard deviation of the input instance $\textbf{X}_r$, respectively. Subsequent denormalization is applied to the prediction outputs to achieve more accurate results. Studies~\cite{ulyanov2017in, he2022masked} indicate that instance normalization effectively mitigates distribution shifts between training and test sets, advantageous for our zero-shot traffic prediction scenarios.

\subsubsection{\bf Patch Embedding for Efficient Long-Term Prediction} 
\label{sec:patch_eb}
% The proposed \model\ model focuses on addressing long-term traffic prediction tasks, which inherently involve processing an increased number of input time steps, leading to significant computational and memory overhead. To mitigate this issue, we employ the patch operation~\cite{dosovitskiy2021an, nie2023a} to partition the data along the temporal dimension. We define $P$ as the patch length, which specifies the number of time steps grouped into a single patch, and $S$ as the stride size, determining the overlap between successive patches. After the patch operation, the input data $\bar{\textbf{X}}_{r}\in\mathbb{R}^{T}$ is reshaped to $\textbf{X}^{P}_{r}\in\mathbb{R}^{P\times N}$, where N is the number of blocks, which can be calculated by $N=\frac{T-P}{S}+1$. 
The proposed \model\ model is designed to address long-term traffic prediction, which often involve processing an increased number of input time steps. This can lead to significant computational and memory overhead. To mitigate these issues, we employ a patch-based approach~\cite{dosovitskiy2021an, nie2023a} to partition the data along the temporal dimension. We define $P$ as the patch length, specifying the number of time steps grouped into a single patch, and $S$ as the stride size, determining the overlap between successive patches. After the patch operation, the input data $\bar{\textbf{X}}_{r}\in\mathbb{R}^{T}$ is reshaped to $\textbf{X}^{P}_{r}\in\mathbb{R}^{P\times N}$, where $N$ is the number of blocks, calculated as $N=\frac{T-P}{S}+1$. Our patch embedding scheme offers several key advantages:
\begin{itemize}[leftmargin=*]

\item \textbf{Handling Temporal Distribution Shift}. By considering one hour of traffic data as the length of a single patch and adjusting the stride accordingly ($S=P$), our model can effectively handle the temporal distribution shifts often encountered in long-term traffic prediction. This allows the model to capture and adapt to the evolving patterns in traffic data over extended time horizons. \\\vspace{-0.12in}

\item \textbf{Computational Efficiency}. The patch-based processing significantly reduces the computational and memory requirements, as the model operates on a smaller number of input patches rather than the entire temporal sequence. This enables more efficient and scalable long-term traffic prediction, making the model suitable for real-world deployments.

\end{itemize}

% Given the varying aggregation frequencies of different traffic data (\eg, every 5 minutes, every 30 minutes), we standardize by considering one hour of traffic data as the length of a single patch and adjusting the stride accordingly ($S=P$). Subsequently, linear transformation and positional encoding are employed to get the spatio-temporal data embedding ${\textbf{E}}_{r}\in\mathbb{R}^{P\times d}$, as follows:

After the patch-based partitioning of the input data, we employ a linear transformation and positional encoding to obtain the final spatio-temporal data embedding ${\textbf{E}}_{r}\in\mathbb{R}^{P\times d}$. This embedding serves as the input to the subsequent model components:
\begin{align}
    \label{eq:Patch_embeddin}
    {\textbf{E}}_{r} = W_e \cdot \textbf{X}^{P}_{r} + \textbf{PE}(\textbf{X}^{P}_{r})
\end{align}
Here, $W_e\in\mathbb{R}^{N\times d}$ is the weight parameter, and $\textbf{PE}$ denotes the Sinusoidal Position Encoding~\cite{vaswani2017attention}, which helps the model capture temporal relationships within each patch. Our temporal patch embedding schema compresses the input time steps, enabling the model to efficiently perform long-term traffic prediction and overcome challenges posed by increased input time steps, which is crucial for accurate long-term forecasting.

\vspace{-0.05in}
\subsection{Spatio-Temporal Context Encoding}
To capture the complex spatio-temporal patterns inherent in traffic data, our model integrates both temporal and spatial context cues. By explicitly modeling the interplay between these two critical dimensions, \model\ is able to better understand the multifaceted factors influencing transportation patterns. This integrated approach enables our framework to generate more accurate forecasts across a diverse range of time horizons and geographic areas.

\label{sec:STC_encding}
\subsubsection{\bf Temporal Context Encoding}
% Traffic data often exhibits periodic patterns, such as the morning and evening rush hours during commuting times. To effectively capture this characteristic, we incorporate temporal information into our model. Specifically, we utilize the patch operation to segment the temporal context, including the time of day $z^{(d)} \in \mathbb{R}^T$ and day of the week $z^{(w)} \in \mathbb{R}^T$. After that, the linear layers are used to generate time embeddings, formalized as follows:
% Traffic data often exhibits distinct temporal patterns, such as periodic variations caused by daily or weekly routines, as well as more complex non-linear trends and dependencies over longer timescales. To effectively capture these cyclical behaviors, our model incorporates relevant temporal information. Specifically, we leverage the patch-based segmentation to extract features related to the time of day, $z^{(d)} \in \mathbb{R}^T$, and the day of the week, $z^{(w)} \in \mathbb{R}^T$. We then employ linear layers to generate time-specific embeddings that encode these temporal context cues. By explicitly modeling the periodic nature of traffic flow, our approach is well-equipped to make accurate predictions, even for long-term forecasting horizons where these temporal patterns become increasingly relevant. The time embeddings are formalized:
Our model effectively captures the distinct temporal patterns in traffic data, such as periodic variations caused by daily or weekly routines, as well as complex non-linear dependencies over longer timescales. Specifically, we leverage patch-based segmentation to extract features related to the time of day, $z^{(d)} \in \mathbb{R}^T$, and the day of the week, $z^{(w)} \in \mathbb{R}^T$, and employ linear layers to generate time-specific embeddings that encode these temporal context cues. By explicitly modeling the periodic nature of traffic flow, our approach is well-equipped to make accurate predictions, even for long-term forecasting horizons where these temporal patterns become increasingly relevant.
\begin{align}
    \label{eq:TE}
    \textbf{D} = \text{concat}[W_1 z^{(d)}_P, W_2 z^{(w)}_P]
\end{align}
% Here, $z^{(d)}_P$, $z^{(w)}_P\in \mathbb{R}^{P\times N}$ denote the temporal context after patching. $W_1$, $W_2\in \mathbb{R}^{N\times d/2}$ are the weight parameters. We concatenate the temporal context embeddings to obtain the final temporal embeddings $\textbf{D} \in \mathbb{R}^{T\times d}$ in our model.
% The temporal context is first extracted as $z^{(d)}_P, z^{(w)}_P \in \mathbb{R}^{P\times N}$ after the patch-based segmentation. We then use linear layers parameterized by $W_1, W_2 \in \mathbb{R}^{N\times d/2}$ to generate the time-of-day and day-of-week embeddings, respectively. Finally, we concatenate these temporal embeddings to obtain the final temporal encoding $\textbf{D} \in \mathbb{R}^{P\times d}$, which is integrated into our model.
The temporal context is extracted as $z^{(d)}_P, z^{(w)}_P \in \mathbb{R}^{P\times N}$ after patch-based segmentation. We use linear layers parameterized by $W_1, W_2 \in \mathbb{R}^{N\times d/2}$ to generate time-of-day and day-of-week embeddings, which are then concatenated into the final temporal encoding $\textbf{D} \in \mathbb{R}^{P\times d}$ and integrated into our model.

\subsubsection{\bf Spatial Context Encoding}
Traffic patterns vary across regions due to their unique geographical characteristics, such as the higher volumes experienced at transportation hubs. To capture these regional attributes, we incorporate the underlying spatial context within the traffic network. First, we calculate the normalized Laplacian matrix $\bm{\triangle} = \bf{I - D^{-1/2}AD^{-1/2}}$, where $\textbf{I}$ and $\textbf{D}$ are the identity and degree matrices, respectively. We then perform an eigenvalue decomposition, obtaining $\bm{\triangle} = \bf{U\Lambda U^\mathsf{T}}$, with $\bf U$ and $\bf \Lambda$ as the eigenvalue and eigenvector matrices. The $k$ smallest non-trivial eigenvectors are used as the region embeddings $\Phi \in \mathbb{R}^{R \times k}$, encoding the structural information of the traffic network. These embeddings are then processed through a linear layer to obtain the final spatial encodings $\textbf{C}\in \mathbb{R}^{R\times d}$.

% Different regions exhibit variations in transportation modes due to their distinct geographical characteristics. For example, transportation hubs experience a higher volume of traffic. To model this attribute, we introduce the traffic network structure to encode regional information. We begin by calculating the normalized Laplacian matrix through $\bm{\triangle} = \bf{I - D^{-1/2}AD^{-1/2}}$, where $\textbf{I}$, $\textbf{D}$ denote the identity matrix and the degree matrix respectively. Subsequently, we perform an eigenvalue decomposition of the Laplacian matrix by $\bm{\triangle} = \bf{U\Lambda U^\mathsf{T}}$. Here, $\bf U$ and $\bf \Lambda$ represent the eigenvalue matrix and the eigenvector matrix, respectively. The $k$ smallest non-trivial eigenvectors are used as the region embedding, denote as $\Phi \in \mathbb{R}^{R \times k}$, which preserve the structural information of nodes within the traffic network. The region embeddings are then processed through a linear layer to obtain the final regional embeddings $\textbf{C}\in \mathbb{R}^{R\times d}$.

% \begin{figure}
%     \centering
%     \includegraphics[width=1\columnwidth]{fig/instruction_text.pdf}
%     \vspace{-0.2in}
%     \caption{Illustration of spatio-temporal prompt instructions encoding the time- and location-aware information.}
%     \vspace{-0.1in}
%     \label{fig:instruction}
% \end{figure}

\vspace{-0.05in}
\subsection{\bf Spatio-Temporal Dependencies Modeling}
\label{sec:ST_modeling}

\subsubsection{\bf Temporal Dependencies Modeling}
% To capture the dynamic temporal evolution patterns, we employ the encoder architecture of the Transformer to encode time-dependent relationships. The temporal patterns are encoded from two perspectives: periodic traffic transitions and dynamic traffic patterns.
To highlight the strong generalization capabilities of our foundation model for long-term forecasting, it can effectively handle both periodic and dynamic traffic patterns. Specifically, we build our pre-trained spatio-temporal model upon our developed TimeShift Transformer architecture to encode time-dependent relationships. This allows our model to capture the traffic patterns from two complementary perspectives:

\begin{itemize}[leftmargin=*]
    \item \textbf{Periodic Traffic Transitions}. Our model captures the periodic, recurring traffic patterns, such as hourly, daily, and weekly cycles. By encoding these cyclical variations, our approach can better account for the inherent regularities in transportation network.
    \item \textbf{Dynamic Traffic Patterns}. Beyond just periodic variations, our temporal encoder also captures the complex, non-linear temporal dynamics and evolving trends in the traffic data over time.
\end{itemize}

Our model's ability to encode both the periodic and dynamic aspects of temporal information enables strong generalization capabilities, making it well-suited for real-world traffic forecasting across diverse time horizons. This versatile understanding of the multifaceted factors influencing transportation patterns allows accurate predictions of not only regular, recurring traffic behaviors, but also unpredictable, evolving changes in conditions. \\\vspace{-0.12in}

% \noindent \textbf{Periodic Association Encoding:} We utilize temporal embeddings $\mathbf{D}$ and spatial embeddings $\mathbf{C}$ to learn periodic time series patterns. Specifically, our goal is to explore the correlations between traffic patterns at historical moments and future instances. To this end, the temporal embeddings are updated to $\mathbf{D}^{his}$ and $\mathbf{D}^{pre} \in \mathbb{R}^{R \times P \times d}$, where "his" denotes the historical temporal information, and "pre" indicates the temporal information anticipated for the future.
\noindent \textbf{Modeling Periodic Traffic Transitions}. To learn the periodic patterns in the traffic, we leverage both temporal embeddings $\mathbf{D}$ and spatial embeddings $\mathbf{C}$. Our objective is to uncover the correlations between historical traffic patterns and future instances. Specifically, we update the temporal embeddings into two components:
\begin{itemize}[leftmargin=*]

\item $\mathbf{D}^{his} \in \mathbb{R}^{R \times P \times d}$: This captures the historical temporal signals.

\item $\mathbf{D}^{pre} \in \mathbb{R}^{R \times P \times d}$: This represents the anticipated temporal information for future predictions.

\end{itemize}
Our model explicitly models both historical and future-oriented temporal patterns, allowing it to better learn and leverage the periodic nature of the traffic time series.

% To achieve dimension alignment, We employ repetitive operations to extend the dimensions of the updated temporal embeddings in the spatial domain, and similarly, the dimensions of spatial embeddings are also extended along the temporal axis. Subsequently, future spatio-temporal embeddings, historical spatio-temporal embeddings, and historical spatio-temporal data representations serve as the query, key, and value, respectively, in the multi-head attention mechanism. The process is formalized as follows:
% To ensure proper dimension alignment, we employ repetitive operations to extend the temporal embeddings along the spatial domain, and likewise, we extend the spatial embeddings across the temporal axis. With these extended embeddings, we then construct the inputs to a multi-head attention mechanism: i) The future spatio-temporal embeddings serve as the query. ii) The historical spatio-temporal embeddings act as the key. iii) The historical spatio-temporal data representations function as the value. The formal definition of this process is as follows:
To ensure dimension alignment, we extend the temporal embeddings along the spatial domain, and the spatial embeddings across the temporal axis. We then construct the inputs to a multi-head attention mechanism: the future spatio-temporal embeddings as the query, the historical spatio-temporal embeddings as the key, and the historical spatio-temporal data representations as the value.
\begin{align}
    \label{eq:QKV}
    Q_r^h = W_q^h (\mathbf{D}^{pre}_r + \bar{\textbf{C}}_r);~~~ K_r^h = W_k^h (\mathbf{D}^{his}_r + \bar{\textbf{C}}_r);~~~ V_r^h = W_k^h {\textbf{E}}_{r}
\end{align}

\begin{align}
    \label{eq:ATT}
    \textbf{M}_r^h = Attention(Q_r^h, K_r^h, V_r^h) = \text{softmax}(\frac{\delta_a(Q_r^h K_r^{h^\mathsf{T}})}{\sqrt{d_h}}) V_r^h
\end{align}
% \noindent Here, $W_q^h$, $W_k^h$ and $W_v^h\in \mathbb{R}^{d \times d_h}$ are weight parameters and $\delta_a$ is the dropout operation. $\bar{\textbf{C}}_r \in \mathbb{R}^{P \times d}$ is the updated spatial embedding and $\textbf{M}_r^h \in \mathbb{R}^{P \times d_h}$ is the output of periodic association encoding from h-th head. After that, RMSNorm~\cite{} is incorporated to enhance the training stability of the model. The aforementioned process explicitly models the relationship between historical and future temporal information in traffic scenarios, effectively facilitating the model in uncovering periodic spatio-temporal patterns.
\noindent Here, $W_q^h$, $W_k^h$, and $W_v^h$ are learnable weight parameters of size $\mathbb{R}^{d \times d_h}$, and $\delta_a$ is a dropout operation. $\bar{\textbf{C}}_r \in \mathbb{R}^{P \times d}$ is the updated spatial embedding, and $\textbf{M}_r^h \in \mathbb{R}^{P \times d_h}$ is the output of the periodic association encoding from the h-th attention head. We incorporate RMSNorm~\cite{zhang2019root} to improve training stability, explicitly modeling the relationship between historical and future temporal information to enable the discovery of periodic spatio-temporal traffic patterns.\\\vspace{-0.12in}

\noindent \textbf{Learning with Dynamic Traffic Patterns}. This module designed to capture the dynamic dependencies among different time segments. For instance, events such as sudden traffic accidents may lead to a sharp decrease in traffic speed and traffic volume. Thus, we empower \model\ to model these types of dynamic relationships, which are crucial for understanding and predicting traffic patterns. To achieve this, we employ an attention mechanism similar to the periodic traffic transition encoding. The difference is that the query (Q), key (K), and value (V) inputs are replaced by the normalized output (M) from the previous step. This modification allows the model to focus on learning the dynamic dependencies between different time segments, rather than just the periodic patterns. The resulting temporal representation $\textbf{H}\in \mathbb{R}^{R\times P \times d}$ captures these dynamic spatio-temporal relationships.

% Dynamic temporal association focuses on capturing the dynamic dependencies among different time segments. For instance, events such as sudden traffic accidents may lead to a sharp decrease in traffic speed. We employ an attention mechanism to model these relationships. The computational process is similar to the periodic association encoding, with the difference being that the $Q$, $K$, and $V$ are replaced by $\textbf{M}$ after normalization. Subsequently, we obtain the temporal representation $\textbf{H}\in \mathbb{R}^{R\times P \times d}$.
\vspace{-0.05in}
\subsubsection{\bf Spatial Dependencies Modeling}
% Spatial dependency association is a crucial characteristic of transportation networks, where the traffic conditions in one region are often influenced by the traffic states of neighboring regions. In our \model\ model, we use graph convolutional networks to model the spatial associations between regions, formalized as follows:
Capturing spatial dependencies is a crucial aspect of our model design, as transportation networks exhibit strong spatial correlations where the traffic conditions in one region are often influenced by the states of neighboring areas. To model these spatial associations, we employ graph convolutional networks (GCNs). The formalization is as follows:
\begin{align}
    \label{eq:GCN}
    \textbf{G}_t = \delta[\alpha \textbf{H}_t + (1-\alpha)(W_g \bar{\textbf{A}} \textbf{H}_t)]
\end{align}
% In this context, $\bar{\textbf{A}}$ represents the normalized weights of the adjacency matrix, and $W_g\in \mathbb{R}^{d\times d}$ denotes learnable parameters. We use $\alpha$ to balance the preservation of original information and utilize $\delta$ to mitigate overfitting. Residual networks, RMSNorm (RN), and SwiGLU~\cite{shazeer2020glu} are employed for subsequent computations. Here, we utilize SwiGLU to replace the feed-forward neural network, which has been widely used and validated within large language model frameworks. The process can be formalized as follows:
% Here, $\bar{\textbf{A}}$ represents the normalized weights of the adjacency matrix, and $W_g\in \mathbb{R}^{d\times d}$ denotes learnable parameters. The $\alpha$ parameter allows us to balance the preservation of original information and the $\delta$ function helps mitigate overfitting. For subsequent computations, we leverage residual connections, RMSNorm (RN), and the SwiGLU~\cite{shazeer2020glu} activation, which has been successfully employed in large language models. The SwiGLU operation replaces the standard feed-forward neural network:
Here, $\bar{\textbf{A}}$ is the normalized adjacency matrix weights, and $W_g\in \mathbb{R}^{d\times d}$ are learnable parameters. The $\alpha$ parameter balances original information preservation, and the $\delta$ function helps mitigate overfitting. For subsequent computations, we use residual connections, RMSNorm (RN), and the SwiGLU~\cite{shazeer2020glu} activation, which has been successful in large language models.
\begin{align}
    \label{eq:FFN}
    \textbf{O}_{r,t}^{(l)} = \text{SwiGLU}[\text{RN}(\textbf{G}_{r,t}^{(l)}+\textbf{O}_{r,t}^{(l-1)})] + \textbf{G}_{r,t}^{(l)}
\end{align}

\begin{align}
    \label{eq:SwiGLU}
    \text{SwiGLU}(\textbf{E}_{r, t}) = W_c [\sigma(W_a \textbf{E}_{r, t}) \cdot W_b \textbf{E}_{r, t}]
\end{align}
% We regard the aforementioned spatio-temporal dependency modeling process as the spatio-temporal encoding network, where $\textbf{O}_{r,t}^{(l)}\in \mathbb{R}^{d}$ denotes the final output of $l$-th layer spatio-temporal encoding network. $\sigma$ represents the Swish activation function, and $W_a, W_b, W_c \in \mathbb{R}^{d\times d}$ are trainable parameter matrices. The proposed \model\ model captures complex spatio-temporal dependencies by stacking $L$ layers of spatio-temporal encoding networks.
Here, $\textbf{O}_{r,t}^{(l)}\in \mathbb{R}^{d}$ represents the final output of the $l$-th layer spatio-temporal encoding network, $\sigma$ is the Swish activation function, and $W_a, W_b, W_c \in \mathbb{R}^{d\times d}$ are trainable parameters. The proposed model captures complex spatio-temporal dependencies by stacking multiple layers of this spatio-temporal encoding network, allowing it to learn the intricate relationships within transportation networks.

\section{Evaluation}
\label{sec:eval}
In our experimental evaluation, we address the following six key research questions to validate the capabilities of our model:
\begin{itemize}[leftmargin=*]
% \item \textbf{RQ1}: Can the proposed \model\ model effectively generalize to zero-shot traffic prediction scenarios?
\item \textbf{RQ1}: Can the proposed \model\ model effectively generalize to zero-shot and long-term traffic forecasting scenarios? \\\vspace{-0.1in}
% \item \textbf{RQ2}: How does the proposed one-for-all traffic foundation model perform in a supervised setting compared to existing baselines?
\item \textbf{RQ2}: How effectively does the proposed \model\ framework integrate cross-data traffic patterns into a single and coherent model in supervised learning environments?\\\vspace{-0.1in}
% \item \textbf{RQ3}: How do different modules contribute to the performance gains of \model?
\item \textbf{RQ3}: Does the model have the capability to rapidly generalize to new transportation forecasting demands as they emerge? \\\vspace{-0.1in}
\item \textbf{RQ4}: What are the individual contributions of the different modules to the performance gains of the \model\ approach?\\\vspace{-0.1in}
% \item \textbf{RQ4}: How do the amount of parameters and data volume affect model performance?
\item \textbf{RQ5}: How the model's parameter count and training data volume scale to affect its overall forecasting performance?\\\vspace{-0.1in}
% \item \textbf{RQ5}: Does \model\ have the capability to rapidly generalize new transportation forecasting demands emerge?
% \item \textbf{RQ6}: Compared to existing large-scale spatio-temporal prediction models, what advantages does our model have in terms of prediction performance and efficiency?
\item \textbf{RQ6}: Compared to existing large-scale spatio-temporal prediction models, what key advantages does our approach offer in terms of prediction performance and efficiency?
\end{itemize}

\subsection{Experimental Setup}
\subsubsection{\bf Data Sources and Characteristics}
The model's generalization capabilities and predictive performance were extensively evaluated using a diverse set of large-scale, real-world public datasets covering various traffic-related data categories, including \textbf{Traffic Flow}, \textbf{Taxi Demand}, \textbf{Bicycle Trajectories}, \textbf{Traffic Speed Statistics}, and \textbf{Traffic Index Statistics}, from regions across the United States and China, such as New York City, Chicago, Los Angeles, the Bay Area, Shanghai, Shenzhen, and Chengdu. 
In addition, we released 3 versions \model\ based on parameter size: $\text{\model}_\text{mini}$ (2M), $\text{\model}_\text{base}$ (5M), and $\text{\model}_\text{plus}$ (26M).
Further details on all experimental datasets are provided in the Appendix.
In the pre-training stage of the \model\ approach, we leveraged a wealth of transportation data, encompassing the traffic flow, traffic speed, and taxi demand datasets. This expansive dataset covers a total of 10,110 regions and 352,796 time points, amounting to an astounding 151,089,924 observations. For the testing phase, we selected data that was outside the training set, allowing us to assess the model's generalization performance across a diverse range of traffic prediction scenarios. These scenarios include:
\begin{itemize}[leftmargin=*]

\item \textbf{Cross-Region Zero-Shot Evaluation}: Assessing the \model's ability and robustness to generalize to unseen regions within a city, without the need for additional training. \\\vspace{-0.1in}

\item \textbf{Cross-City Zero-Shot Evaluation}: Examining the model's capacity to adapt to completely new cities, leveraging the knowledge acquired during the original training process. \\\vspace{-0.1in}

\item \textbf{Cross-Task Zero-Shot Evaluation}: Testing the model's ability to forecast different types of traffic-related data, including traffic flow, speed, and demand, without any additional training. \\\vspace{-0.1in}

\item \textbf{Unified Model Supervised Evaluation}: Evaluating the versatility and adaptability of our unified \model\ within a supervised learning framework, focusing on its ability to handle diverse spatio-temporal traffic scenarios. \\\vspace{-0.1in}

\item \textbf{Cross-Data Fast Adaptation Evaluation}: Measuring the model's cost-efficient adaptation capabilities to new traffic datasets while requiring only a small amount of additional training data, in contrast to the need for extensive retraining from scratch.

\end{itemize}

\begin{table*}[t]
\renewcommand\arraystretch{0.97}
    \centering
    \small
    \caption{Comparison of performance between \model\ in zero-shot setting and the baselines in full-shot setting.}
    \vspace{-0.15in}
    \label{tab:zero-shot}
    \scalebox{0.98}{
    \begin{tabular}{c c c c c c c c c c c c c}
        \hline
        % \multirow{2}{*}{Dataset} & \multirow{2}{*}{Dataset} & \multicolumn{3}{c}{Zero-shot} & \multicolumn{10}{c}{Full-shot} \\
        % \cline{2-15}
        % & \multicolumn{1}{c}{\model-10M} & \multicolumn{1}{c}{\model-100M} & \multicolumn{1}{c}{\model-0.1B} & \multicolumn{1}{c}{TGCN} & \multicolumn{1}{c}{STGCN} & \multicolumn{1}{c}{ASTGNN} & \multicolumn{1}{c}{GWN} & \multicolumn{1}{c}{STSGCN} & \multicolumn{1}{c}{MTGNN} & \multicolumn{1}{c}{AGCRN} & \multicolumn{1}{c}{MSDR} & \multicolumn{1}{c}{STWA} & \multicolumn{1}{c}{PDFormer}\\
        % \cline{2-15}

        \multirow{2}{*}{Dataset} & \multirow{2}{*}{Evaluation} & \multicolumn{1}{c}{\textbf{Zero-shot}} & \multicolumn{10}{c}{\textbf{Full-shot}} \\
        \cmidrule(lr){3-3}\cmidrule(lr){4-13}
        & & $\textbf{\model}_\text{plus}$ & PDFormer & STWA & MSDR & AGCRN & MTGNN & STSGCN & GWN & ASTGNN & STGCN & TGCN \\
        \cline{1-13}
        \multirow{3}{*}{CAD3} 
        & MAE & \textbf{15.88} & 20.28 & 21.65 & 22.74 & 18.72 & 17.59 & 21.88 & \underline{16.94} & 23.60 & 20.24 & 19.56 \\
        & RMSE & \textbf{27.03} & 36.43 & 37.55 & 37.15 & 31.93 & 28.92 & 34.52 & \underline{28.81} & 39.35 & 34.34 & 30.82 \\
        & MAPE & \textbf{21.94\%} & 25.19\% & 26.85\% & 32.37\% & 25.45\% & 25.22\% & 30.20\% & \underline{22.98\%} & 41.22\% & 25.33\% & 28.15\% \\
        \cline{1-13}
        \multirow{3}{*}{CAD5} 
        & MAE & \underline{11.09} & 12.89 & 14.43 & 15.44 & 12.88 & 11.70 & 13.87 & \textbf{10.69} & 12.58 & 13.76 & 13.07 \\
        & RMSE & \textbf{18.96} & 21.18 & 24.14 & 29.52 & 23.78 & 20.30 & 22.32 & \underline{19.75} & 21.23 & 27.49 & 21.56 \\
        & MAPE  & 27.90\% & 28.82\% & 29.34\% & 32.49\% & \underline{26.29\%} & 26.75\% & 32.84\% & \textbf{25.98\%} & 30.56\% & 32.89\% & 32.65\% \\
        \cline{1-13}
        \multirow{3}{*}{PEMS07M} 
        & MAE & 4.50 & 4.62 & 4.54 & 4.81 & 4.61 & 4.52 & 4.56 & \textbf{4.17} & \underline{4.39} & 4.44 & 4.88 \\
        & RMSE & 8.21 & 8.36 & 8.57 & 8.61 & 8.63 & 8.06 & \underline{8.05} & \textbf{7.84} & 8.21 & 8.37 & 8.38 \\
        & MAPE & \underline{12.20\%} & 13.74\% & 12.91\% & 13.90\% & 12.81\% & 13.11\% & 12.70\% & \textbf{11.46\%} & 12.58\% & 12.59\% & 13.94\% \\
        \cline{1-13}
        \multirow{3}{*}{TrafficSH} 
        & MAE & \textbf{0.55} & 0.77 & 1.42 & 1.84 & 1.33 & 0.81 & 1.66 & 0.76 & \underline{0.69} & 1.60 & 1.79 \\
        & RMSE & \textbf{0.85} & 1.23 & 2.49 & 3.55 & 2.45 & 1.26 & 3.33 & 1.39 & \underline{1.09} & 3.14 & 2.65 \\
        & MAPE & \textbf{8.01\%} & 8.34\% & 9.92\% & 9.91\% & 8.51\% & 8.29\% & 9.33\% & 9.23\% & 8.05\% & \underline{8.04\%} & 17.75\% \\
        \cline{1-13}
        \multirow{3}{*}{CHI-TAXI} 
        & MAE  & \textbf{1.91} & 4.03 & 3.70 & 3.55 & 3.60 & 3.27 & 4.87 & 3.56 & 3.28 & \underline{3.09} & 4.02 \\
        & RMSE  & \textbf{4.42} & 12.82 & 11.49 & 10.39 & 11.31 & 9.87 & 14.40 & 11.27 & 10.32 & \underline{9.54} & 11.70 \\
        & MAPE  & \underline{40.07\%} & 44.42\% & 42.52\% & 52.88\% & 46.48\% & \textbf{39.38\%} & 104.64\% & 41.31\% & 42.82\% & 42.47\% & 60.25\% \\
        \cline{1-13}
        \multirow{3}{*}{NYC-BIKE} 
        & MAE & \textbf{6.32} & 7.33 & 7.97 & 8.08 & 8.13 & 6.48 & 6.85 & 7.61 & \underline{6.44} & 8.01 & 7.75 \\
        & RMSE & \underline{11.60} & 13.01 & 14.35 & 13.97 & 14.46 & \textbf{11.49} & 11.98 & 13.56 & 11.62 & 13.94 & 13.49 \\
        & MAPE & 70.06\% & 65.44\% & 64.19\% & 77.56\% & 78.36\% & \underline{61.52\%} & 68.44\% & \textbf{58.41\%} & 63.90\% & 63.02\% & 85.27\% \\
        \hline
        % % \hline
    \end{tabular}
    }
    \vspace{-0.05in}
\end{table*}

% We released three different versions of the \model\ model with varying amounts of parameters: $\textbf{\model}_\text{mini}$, $\textbf{\model}_\text{base}$, and $\textbf{\model}_\text{plus}$, with \textbf{2M}, \textbf{5M}, and \textbf{27M} parameters respectively. The model scale is expanded by increasing the dimensions of the hidden layers and the number of layers in the spatio-temporal encoder, as detailed in Table 1. Given our focus on long-term traffic forecasting, the time spans for both historical timesteps $H$ and future timesteps $F$ are set to 1 day, determined by the aggregation frequency of different datasets (\eg, $H=F=288$ when the aggregation frequency is 5 minutes). Similarly, the patch length $P$ and stride $S$ are both set to one-hour time spans, which correspond to a value of 12 when the aggregation frequency is 5 minutes. The regional embedding-related hyperparameter $k$ is set to 8, and the balancing weight $\alpha$ is set to 0.05. The dropout ratios for the attention matrix $\delta_a$ and the spatio-temporal network $\delta$ are set to 0.3 and 0.1, respectively. We maximize the batch size settings based on the GPU memory usage across different models. Moreover, the hyperparameters for all baseline configurations adhere to the settings provided in the original papers or the official released codes. Training and testing of all models are conducted on a server equipped with 8 $\times$ NVIDIA A100-SXM4-40GB GPUs. More details on the hyperparameter settings are provided in the supplementary materials.
% \vspace{-0.05in}
\subsection{Zero-shot vs. Full-shot Performance (RQ1)}
In this section, we focus on evaluating the zero-shot generalization capability of the \model. After its pre-training phase, the \model\ is directly applied to zero-shot prediction tasks on various downstream datasets, including \textbf{Cross-Region}, \textbf{Cross-City}, and \textbf{Cross-Category} settings. In contrast, the baseline models are developed under a supervised learning framework, where they undergo training on these downstream datasets before being tested. The experimental setups are labeled as \textbf{Zero-shot} for the \model and \textbf{Full-shot} for the baselines, with the results displayed in Table 2. The best performance is highlighted in bold, while the second-best performance is indicated with an underline. \\\vspace{-0.12in}

% In this section, we focus on evaluating the zero-shot generalization capability of \model. After its pre-training phase, \model\ is directly applied to zero-shot prediction tasks on various downstream datasets, which include both cross-region and cross-category settings. The term 'cross-region' refers to testing in untrained geographical regions, while 'cross-category' denotes testing on previously unseen types of traffic data. In contrast, the baselins are developed under a supervised learning framework, where they undergo training on these downstream datasets before being tested. The experimental setups are labeled as \textbf{Zero-shot} for \model\ and \textbf{Full-shot} for the baselines, with the results displayed in Table 2. Here, bold text highlights the best performance, while an underline indicates the second-best performance. Based on the results, we can draw three key observations:

\noindent \textbf{(i) Outstanding Zero-shot Prediction Performance.}
\model\ achieves significant zero-shot learning breakthroughs, outperforming most baselines even without fine-tuning. This highlights the approach's robustness and effectiveness at learning complex spatio-temporal patterns in large-scale traffic data, extracting universal insights applicable across downstream tasks.

\model\ consistently secures top or second positions on several datasets, and maintains a competitive performance gap within 8\% MAE even when not leading. This outstanding zero-shot prediction performance underscores the \model's versatility and adaptability in handling diverse traffic datasets without extensive retraining. A crucial advantage is its readiness for immediate deployment in new scenarios, significantly reducing the time and resources typically required by traditional supervised approaches, offering substantial benefits for practical applications. \\\vspace{-0.12in}

\noindent \textbf{(ii) Exceptional Cross-Task Generalization}.
Our model was evaluated across four distinct traffic data categories: traffic flow (CAD3, CAD5), traffic speed (PEMS07M, TrafficSH), taxi demand (CHI-TAXI), and bicycle trajectories (NYC-BIKE). The baseline analysis revealed that while various models performed exceptionally well on specific data types, none could consistently deliver top-tier results across all categories. For instance, GWN, STGCN, and ASTGCN each exhibited remarkable capabilities in predicting traffic flow (CAD3, CAD5), taxi demand (CHI-TAXI), and traffic speed (TrafficSH) respectively. However, they struggled to maintain that level of performance in other domains. In contrast, the \model\ consistently delivered high-quality results across all tested categories, underscoring its exceptional robustness and versatility.

To assess the versatility of our \model\ framework, we evaluated its cross-category zero-shot generalization during testing. This involved incorporating bicycle trajectory data (NYC-BIKE), despite its absence in pretraining. The results were highly promising, as \model\ maintained excellent performance on both MAE and RMSE metrics, further validating its universality and ability to adapt across diverse data types in real-life urban scenarios. \\\vspace{-0.12in}

% \textbf{(iii) Exceptional Cross-Task Generalization}
% During the testing phase, we incorporated bicycle trajectory data (NYC-BIKE) to evaluate \model's cross-category zero-shot generalization ability, despite this traffic feature being absent during the pretraining stage. The results show that \model\ maintained excellent performance on both the MAE and RMSE metrics in this scenario, further validating the universality of the proposed framework. By adeptly extracting both periodic and dynamic spatio-temporal patterns from varied traffic data distributions, \model\ has demonstrated its ability to adapt and perform reliably across different data types, significantly boosting its cross-category prediction capabilities.

\noindent \textbf{(iii) Strong Long-term Forecasting Capabilities}. A key strength of our \model\ architecture is its exceptional temporal generalization ability, which enables it to outperform baseline methods in long-term traffic prediction tasks. Many existing models often struggle to maintain accurate forecasts over extended time horizons, as they tend to overfit to historical patterns and fail to adequately capture the dynamic and evolving nature of traffic conditions. In contrast, \model\ has demonstrated a remarkable capacity to learn universal spatio-temporal representations from diverse traffic data sources. This allows the model to generate robust predictions that remain reliable even as traffic patterns shift and evolve over time.

\begin{table*}[t]
\renewcommand\arraystretch{0.97}
    \centering
    \small
    \caption{Evaluation of predictive performance across different models in a supervised setting.}
    \vspace{-0.15in}
    \label{tab:supervised}
    \scalebox{0.97}{
    \begin{tabular}{c c |c c c | c c c | c c c | c c c | c c c}
        \hline
        \multirow{2}*{Model} & \multicolumn{1}{c}{Dataset} & \multicolumn{3}{c}{CAD8-1} & \multicolumn{3}{c}{CAD8-2} & \multicolumn{3}{c}{CAD12-2} & \multicolumn{3}{c}{PEMS-BAY} & \multicolumn{3}{c}{NYC-TAXI}\\
        \cline{2-17}
        & \multicolumn{1}{c}{Metrics} & MAE & RMSE & MAPE & MAE & RMSE & MAPE & MAE & RMSE & MAPE & MAE & RMSE & MAPE & MAE & RMSE & MAPE\\
        \hline
        \multicolumn{2}{c}{TGCN} & 27.68 & 45.09 & 20.18\% & 23.43 & 37.03 & 15.55\% & 36.53 & 60.01 & 61.60\% & 2.94 & 6.33 & 7.23\% & 6.10 & 12.70 & 80.39\% \\
        \multicolumn{2}{c}{STGCN} & 31.26 & 49.91 & 24.26\% & 24.30 & 39.59 & 18.61\% & 34.60 & 62.47 & 52.49\% & 2.72 & 5.96 & 6.55\% & 4.17 & 9.19 & 45.54\% \\
        \multicolumn{2}{c}{ASTGCN} & 29.38 & 46.84 & 22.51\% & 24.24 & 38.14 & 18.36\% & 35.19 & 57.87 & 59.73\% & 2.82 & 6.31 & 7.06\% & 5.45 & 13.44 & 59.05\% \\
        \multicolumn{2}{c}{GWN} & 29.03 & 49.26 & 23.23\% & 25.17 & 41.47 & 18.50\% & 38.05 & 69.89 & 64.77\% & 2.66 & 5.60 & 5.94\% & 4.28 & 10.59 & 41.82\% \\
        \multicolumn{2}{c}{STSGCN} & 32.38 & 53.28 & 25.63\% & 24.60 & 41.23 & 18.54\% & 37.00 & 63.19 & 58.29\% & 2.87 & 6.10 & 6.80\% & 5.03 & 10.96 & 65.17\% \\
        \multicolumn{2}{c}{MTGNN} & 31.60 & 53.01 & 25.82\% & 21.85 & 34.02 & 14.20\% & 36.20 & 65.55 & 56.86\% & 2.59 & 5.43 & \textbf{5.87\%} & 3.72 & 7.94 & 43.35\% \\
        \multicolumn{2}{c}{AGCRN} & 34.98 & 58.37 & 26.82\% & 24.40 & 39.21 & 16.49\% & 39.91 & 74.47 & 63.36\% & 2.65 & \textbf{5.32} & 6.05\% & 4.55 & 10.11 & 52.45\% \\
        \multicolumn{2}{c}{MSDR} & 28.99 & 46.91 & 23.60\% & 26.67 & 41.50 & 18.96\% & 40.88 & 72.87 & 69.47\% & 2.94 & 5.68 & 6.42\% & 4.22 & 9.08 & 55.50\% \\
        \multicolumn{2}{c}{STWA} & 32.09 & 53.26 & 25.62\% & 26.57 & 44.01 & 20.21\% & 41.21 & 77.78 & 67.72\% & 2.74 & 5.65 & 6.03\% & 6.05 & 15.22 & 54.81\% \\
        \multicolumn{2}{c}{PDFormer} & 29.64 & 49.79 & 20.65\% & 24.32 & 38.95 & 14.84\% & 34.23 & 59.70 & 50.92\% & 2.75 & 6.15 & 6.50\% & 3.64 & 8.24 & 37.40\% \\
        \hline
        \multicolumn{2}{c}{\emph{$\textbf{\model}_\text{plus}$}} 
        & \textbf{22.50} & \textbf{39.06} & \textbf{15.50\%} & \textbf{17.95} & \textbf{29.61} & \textbf{10.97\%} & \textbf{24.20} & \textbf{39.23} & \textbf{33.22\%} & \textbf{2.59} & 5.67 & 5.99\% & \textbf{3.10} & \textbf{6.45} & \textbf{37.39\%} \\        
        \hline
        % \hline
    \end{tabular}
    }
    % \vspace{-0.05in}
\end{table*}

\vspace{-0.10in}
\subsection{Exceptional Supervised Performance (RQ2)}
% In the supervised task evaluation, we directly utilize the pre-trained \model\ model for testing, while other baselines are trained and evaluated on the specific dataset. As illustrated in Table~\ref{tab:supervised}, the results indicate that our \model\ maintains excellent performance in the supervised setting and has a leading advantage in most metrics. In addition, we observed that most baselines underperform on the CAD-X dataset, possibly due to their tendency to overfit historical spatio-temporal patterns, making it challenging to generalize to future prediction scenarios. Our \model\ architecture effectively extracts universal periodic and dynamic spatio-temporal representations from large-scale data, thereby addressing the issue of poor prediction performance caused by temporal distribution shift. \\
% Furthermore, the advantages demonstrated across multiple datasets confirm the proposed \model's \textbf{robust one-for-all generalization capabilities}. This success can be attributed to the architecture's ability to learn universal spatio-temporal knowledge from diverse data sources and then generate robust representations. Such capabilities enhance its generalization potential in downstream tasks, further substantiating the viability of using \model\ as a benchmark model framework.
To further validate the advantage of our \model\ over current baseline methods, we conducted a supervised learning evaluation. In this setting, the \model\ is directly compared to the baseline approaches, which are built in an end-to-end training manner. This supervised evaluation allows us to assess the \model's performance when trained on the target datasets, in contrast to the zero-shot generalization setting previously explored.

As illustrated in Table~\ref{tab:supervised}, the results indicate that our \model\ maintains excellent performance in the supervised setting and holds a leading advantage in most evaluation metrics. Additionally, we observed that most baseline models underperformed on the CAD-X dataset, possibly due to their tendency to overfit historical spatio-temporal patterns, making it challenging for them to generalize to long-term traffic dependency modeling. In contrast, the \model\ architecture effectively extracts universal periodic and dynamic spatio-temporal representations from our pre-training stage, addressing the issue of poor prediction performance caused by cross-time and cross-location distribution shift.

\vspace{-0.1in}
\subsection{Model Fast Adaptation Capabilities (RQ3)}
% This section focuses on evaluating the rapid generalization capabilities of \model\ for downstream tasks. We specifically focus on a type of traffic data previously unseen during \model's pre-training phase-the traffic index data (CD-DIDI and SZ-DIDI). We employ a "Efficient Fine-tuning" approach where only the model's prediction head (the last linear layer) is updated, with a maximum of three training rounds. In contrast, the maximum training and early stopping rounds for other baseline models are set at 100 and 15, respectively, without freezing any parameters. The results are detailed in Table~\ref{tab:fast_adapt}, where "Cost" denotes the total training time in minutes. It is evident that the zero-shot performance of \model\ on some indicators is not as robust as the full-shot performance of the baseline model. This discrepancy may stem from the significant variations in traffic patterns and inconsistent time sampling frequencies between the new and observed data categories. However, following efficient fine-tuning,\model's performance significantly improves, outperforming all compared models, with its training time comprising only 2\%-32\% of that required by the baselines. This enhancement confirms that \model\ can rapidly adapt to new spatio-temporal data categories, underscoring its quick adaptability and further validating its potential as a foundational traffic model.
This section evaluates the swift adaptation abilities of our \model\ for downstream tasks. We focus on a previously unseen traffic dataset and employ an "Efficient Fine-tuning" approach, where only the model's prediction head (the last linear layer) is updated with a maximum of three training epochs. As detailed in Table~\ref{tab:fast_adapt}, the zero-shot performance of \model\ on some indicators is not as robust as the full-shot performance of the baseline models, likely due to variations in traffic patterns and data sampling. However, after efficient fine-tuning, \model's performance substantially improves, outperforming all compared models. Remarkably, the training time for \model\ comprises only 2\%-32\% of that required by the baselines. This rapid adaptability underscores \model's potential as a foundational traffic forecasting model, capable of quickly adapting to new spatio-temporal data categories.

\begin{table}[t]
\renewcommand\arraystretch{0.95}
    \centering
    \small
    \caption{Evaluation of \model's Fast Adaptability.}
    \vspace{-0.15in}
    \label{tab:fast_adapt}
    % \scalebox{1}{
    \scalebox{0.95}{
    \begin{tabular}{l|c c c c c c}
        \hline
        \multirow{2}{*}{Model} & \multicolumn{3}{c}{CD-DIDI} & \multicolumn{3}{c}{SZ-DIDI}\\
        \cline{2-7}
        & \multicolumn{1}{c}{MAE} & \multicolumn{1}{c}{RMSE} & \multicolumn{1}{c}{Cost} & \multicolumn{1}{c}{MAE} & \multicolumn{1}{c}{RMSE} & \multicolumn{1}{c}{Cost}\\
        % \cline{2-9}
        \hline
        STGCN & 3.29 & 4.68 & 38.7 & 2.87 & 4.07 & 46.8 \\
        GWN & 3.33 & 4.75 & 38.8 & 3.14 & 4.47 & 47.8 \\
        MTGNN & 3.38 & 4.80 & 32.38 & 3.08 & 4.32 & 40.4 \\
        AGCRN & 3.55 & 5.03 & 59.4 & 3.16 & 4.54 & 72.0 \\
        STWA & 3.29 & 4.74 & 115.27 & 3.36 & 4.74 & 145.4 \\
        PDFormer & 3.00 & 4.31 & 73.9 & 2.75 & 4.06 & 209.1 \\
        \hline
         & \multicolumn{6}{c}{\textbf{Zero-shot}}\\
        \cline{1-7}
        $\textbf{\model}_\text{mini}$ & 5.05 & 6.81 & - & 3.30 & 4.62 & - \\
        $\textbf{\model}_\text{base}$ & 5.60 & 8.88 & - & 3.71 & 6.15 & - \\
        $\textbf{\model}_\text{plus}$ & 6.03 & 9.50 & - & 3.68 & 5.58 & - \\
        \cline{1-7}
         & \multicolumn{6}{c}{\textbf{Efficient Fine-tuning}}\\
        \cline{1-7}
        $\textbf{\model}_\text{mini}$ & 3.03 & 4.36 & \textbf{2.4} & 2.42 & 3.62 & \textbf{2.8} \\
        $\textbf{\model}_\text{base}$ & \underline{2.99} & \underline{4.30} & \underline{3.3} & \underline{2.40} & \underline{3.59} & \underline{4.1} \\
        $\textbf{\model}_\text{plus}$ & \textbf{2.97} & \textbf{4.29} & 12.3 & \textbf{2.36} & \textbf{3.55} & 14.6 \\
        \hline
    \end{tabular}
    \vspace{-0.3in}
    }
\end{table}

% \vspace{-0.05in}

\begin{figure}[t]
\centering
\subfigure{
    \begin{minipage}[t]{1\linewidth}
        \centering
        \includegraphics[width=2.4in]{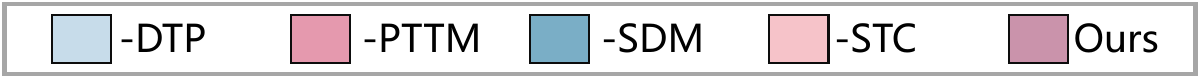}
    \end{minipage}%
}%
\vspace{-0.1in}
\subfigure{
    \vspace{-0.08in}
    \begin{minipage}[t]{0.33\linewidth}
        \centering
        \includegraphics[width=0.9in]{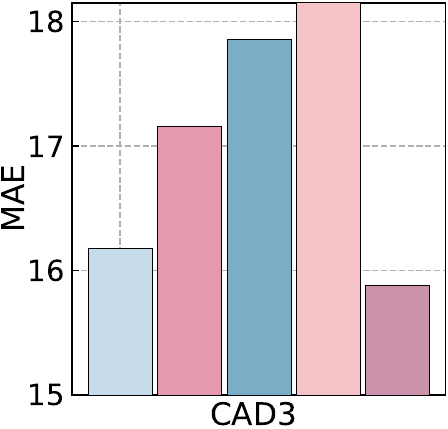}
    \end{minipage}%
    \begin{minipage}[t]{0.33\linewidth}
        \centering
        \includegraphics[width=0.9in]{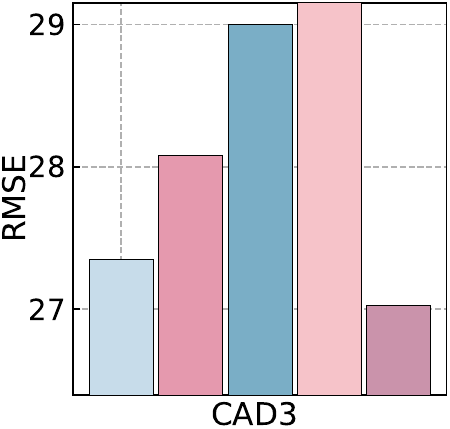}
    \end{minipage}%
    \begin{minipage}[t]{0.33\linewidth}
        \centering
        \includegraphics[width=0.9in]{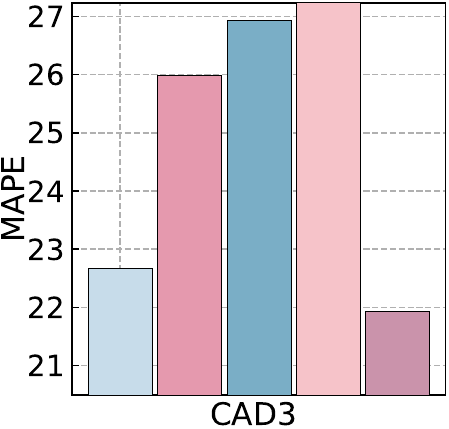}
    \end{minipage}%
}\\\vspace{-0.1in}
\subfigure{
    \begin{minipage}[t]{0.33\linewidth}
        \centering
        \includegraphics[width=0.9in]{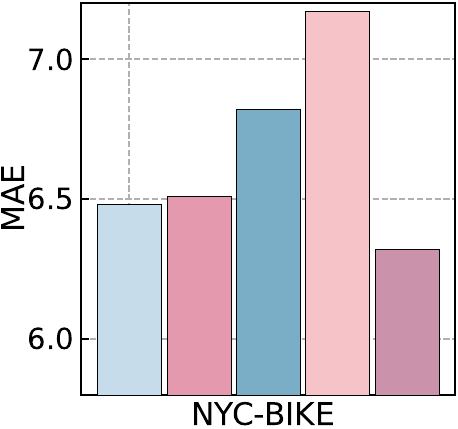}
    \end{minipage}%
    \begin{minipage}[t]{0.33\linewidth}
        \centering
        \includegraphics[width=0.9in]{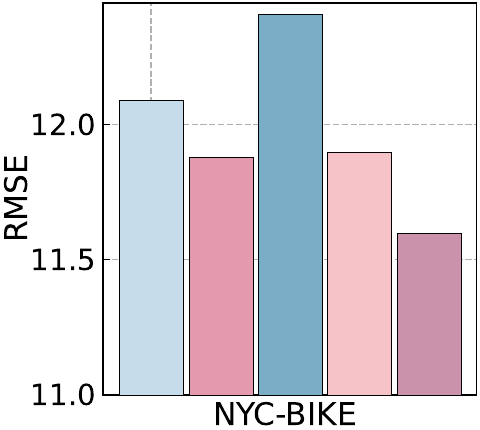}
    \end{minipage}%
    \begin{minipage}[t]{0.33\linewidth}
        \centering
        \includegraphics[width=0.9in]{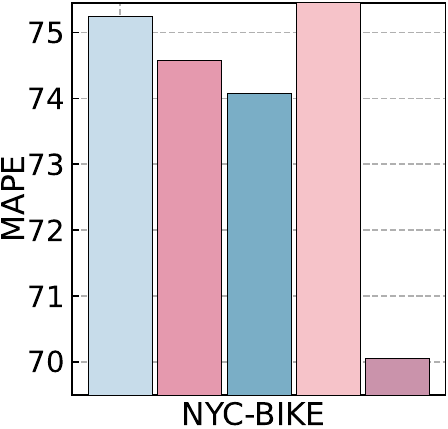}
    \end{minipage}%
}%
\centering
\vspace{-0.4cm}
\caption{Ablation study of our \model\ framework.}
\vspace{-0.4cm}
\label{fig:ablation}
\end{figure}

% \vspace{-0.1in}
\subsection{Ablation Study (RQ4)}
% To assess the individual contributions of the various components within our proposed \model\ model, we conducted ablation studies using the CAD3 and NYC-BIKE datasets in zero-shot scenarios. The results are presented in Figure~\ref{fig:ablation}. This methodological approach involves removing or disabling specific modules within our model and observing the resultant effect on performance metrics, including the follow variants:
To assess the individual contributions of the various components within our proposed \model\ model, we conducted an ablation study using the CAD3 and NYC-BIKE datasets in zero-shot scenarios. The evaluation results are presented in Figure~\ref{fig:ablation}.

\begin{itemize}[leftmargin=*]

\item \textbf{Impact of Learning from Dynamic Traffic Patterns}. In the \emph{-DTP} variant, we removed the Dynamic Traffic Pattern modeling module, which resulted in a decrease in performance. By integrating this encoding process, the model is equipped to thoroughly analyze recent traffic patterns and adapt its predictions effectively in response to sudden changes in traffic conditions. \\\vspace{-0.12in}

\item \textbf{Impact of Periodic Traffic Transition Modeling}.
% -CATT. We excluded the periodic correlation encoding module, integrating temporal and spatial context information directly into the spatio-temporal embedding as additional features. The observed performance degradation indicates that the proposed periodic correlation attention mechanism is more effective in extracting traffic's periodic patterns. By modeling the mapping relationship of traffic flow across various granular historical-future time pairs, the \model\ model successfully captures the general laws governing the evolution of spatio-temporal patterns across different periods, thus facilitating effective zero-shot generalization.\\\vspace{-0.12in}
In the \emph{-PTTM} variant, we excluded the periodic correlation encoding, integrating temporal and spatial context directly into the spatio-temporal embedding. By modeling the mapping of traffic flow across historical-future time pairs, \model\ captures the general laws governing the evolution of spatio-temporal patterns. \\\vspace{-0.12in}

% \item \textbf{Impact of the Spatial Dependencies Modeling.} -GCN. We leave the spatial encoding module unexplored. The analysis reveals that learning spatial relationships significantly enhances spatio-temporal prediction. By aggregating traffic information from adjacent nodes, the model effectively captures dynamic traffic flow patterns, thereby providing a valuable auxiliary signal for zero-shot traffic prediction.\\\vspace{-0.12in}
\item \textbf{Impact of the Spatial Dependencies Modeling.} In the \emph{-SDM} variant, we left the spatial encoding module unexplored. The analysis reveals that learning spatial relationships significantly enhances spatio-temporal prediction. By aggregating traffic information from dependent spatial regions, the model effectively captures dynamic traffic flow patterns, thereby providing a valuable auxiliary signal for zero-shot traffic prediction. \\\vspace{-0.12in}

\item \textbf{Impact of Spatio-temporal Context Encoding.} 
In the \emph{-STC} variant, the encoding of temporal and spatial context information was omitted, resulting in a notable degradation in performance. Temporal context information helps the model identify and learn from common traffic patterns within specific periods, while regional embeddings encapsulate vital area-specific characteristics. These elements collectively provide valuable insights into understanding the dynamic spatio-temporal patterns across cities.

% -STE. The encoding of temporal and spatial context information has been omitted, resulting in the subsequent removal of the associated periodic correlation encoding module. This exclusion leads to a notable degradation in performance, highlighting the critical role that encoding temporal and spatial information plays in spatio-temporal modeling. Temporal data helps the model identify and learn from common traffic patterns within specific periods, thus enhancing its ability to make future predictions. Similarly, regional embeddings encapsulate vital characteristics of different areas, aiding the model in generating more personalized and accurate predictions. These elements collectively provide valuable insights into understanding the dynamic spatio-temporal patterns of various cities.\\\vspace{-0.12in}

\end{itemize}

\vspace{-0.1in}
\subsection{Scaling Law Investigation (RQ5)}
\begin{figure}[t]
\centering
\subfigure{
    \vspace{-0.2in}
        \begin{minipage}[t]{0.25\linewidth}
        \centering
        \includegraphics[width=0.813in]{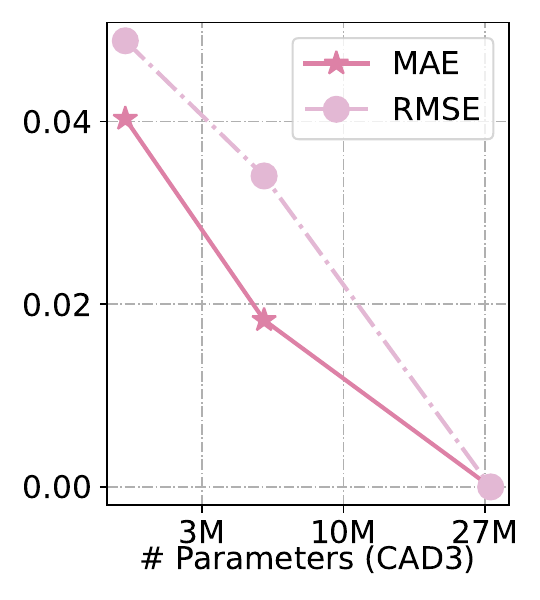}
    \end{minipage}%
    \begin{minipage}[t]{0.25\linewidth}
        \centering
        \includegraphics[width=0.813in]{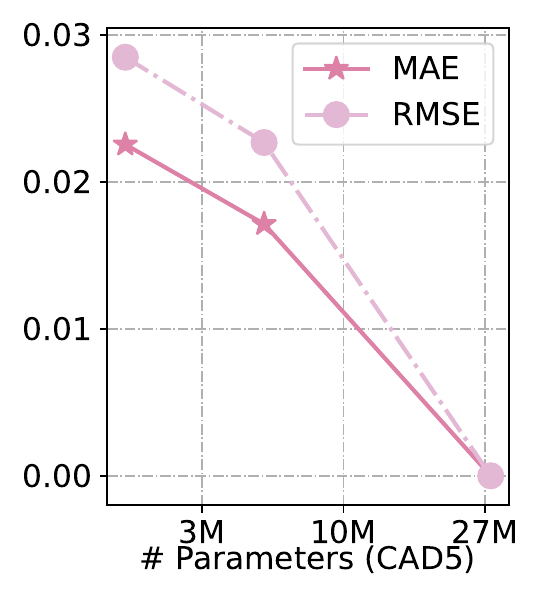}
    \end{minipage}%
    \begin{minipage}[t]{0.25\linewidth}
        \centering
        \includegraphics[width=0.8in]{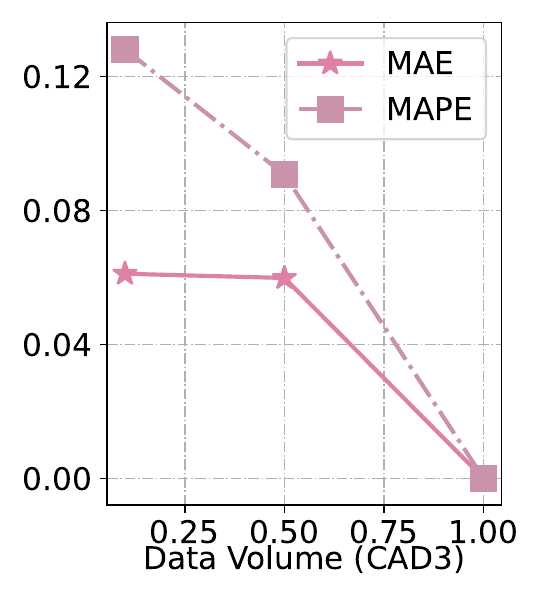}
    \end{minipage}%
    \begin{minipage}[t]{0.25\linewidth}
        \centering
        \includegraphics[width=0.805in]{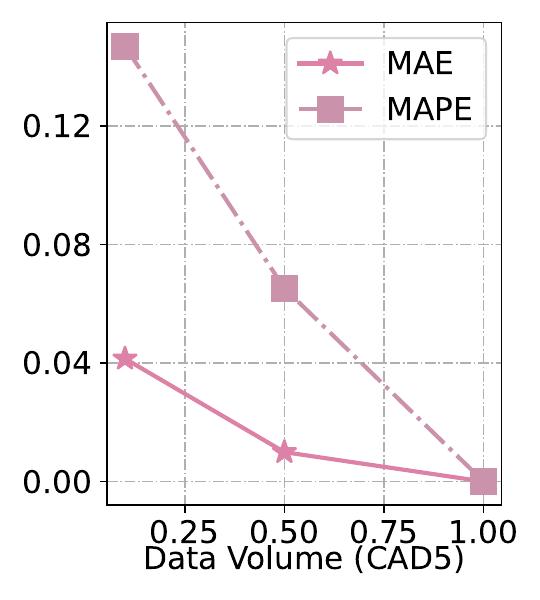}
    \end{minipage}%
}%
\centering
\vspace{-0.30cm}
\caption{Exploration of \model's scalability.}
\vspace{-0.3cm}
\label{fig:scaling}
\end{figure}

% This section explores the scalability of \model, with the findings illustrated in Figure~\ref{fig:scaling}. The investigation into parameter scalability focused on 3 specific versions: $\text{\model}_\text{mini}$ (2M), $\text{\model}_\text{base}$ (5M), and $\text{\model}_\text{plus}$ (27M). Regarding data scalability, we utilized 10\%, 50\%, and 100\% of the pre-training data as training sets for $\text{\model}_\text{plus}$. To standardize comparisons across different metrics, the vertical axis represents relative values that quantify prediction errors. The results indicate that the zero-shot generalization performance of our model progressively improves as both the parameter scale and data scale increase. This improvement suggests that \model\ is capable of extracting valuable knowledge from extensive datasets, and its learning capabilities enhance with the expansion of its parameter base. The demonstrated scalability potential further supports the prospect of \model\ evolving into a foundational model for general transportation applications.
We explored the scalability of \model\ along both data and parameter dimensions, as illustrated in Figure~\ref{fig:scaling}. \emph{Parameter Scalability}: We investigated 3 versions - $\text{\model}_\text{mini}$ (2M), $\text{\model}_\text{base}$ (5M), and $\text{\model}_\text{plus}$ (26M) parameters. \emph{Data Scalability}: For $\text{model}_{plus}$, we utilized 10\%, 50\%, and 100\% of the pre-training data to explore the benefits of incorporating more data. To standardize comparisons, the vertical axis represents relative prediction error values. The results show that \model's zero-shot generalization performance progressively improves as both parameter and data scale increase. This suggests \model's ability to extract valuable knowledge from extensive datasets, with its learning capabilities enhanced by parameter expansion. The demonstrated scalability potential supports \model's prospect as a foundational model for general transportation applications.

\vspace{-0.1in}
\subsection{Comparison with Large Spatio-Temporal Pre-trained Models (RQ6)}
% In this section, we compare \model\ with other prominent large-scale spatio-temporal prediction models, including UniST~\cite{yuan2024unist} and UrbanGPT~\cite{li2024urbangpt}, both known for their good zero-shot generalization capabilities. Specifically, we utilized the CHI-TAXI dataset, which was not included in the pre-training phase for any of the three models, for our evaluation. Detailed settings are provided in the Appendix. The results presented in Table~\ref{tab:compare_large} illustrate that \model\ maintains a significant performance advantage over other advanced large-scale spatio-temporal models. In addition, it is evident that \model\ and UniST exhibit significant improvements in efficiency compared to UrbanGPT. This enhancement may be due to UrbanGPT's reliance on the LLM for making predictions through a question-and-answer format, which hinders its ability to process batch data efficiently. The proposed \model\ model achieves a win-win in performance and efficiency, highlighting its potential as a large model for traffic benchmarks.
In this section, we compare our proposed \model\ with other prominent large spatio-temporal pre-trained models, including UniST~\cite{yuan2024unist} and UrbanGPT~\cite{li2024urbangpt}, known for their strong zero-shot generalization. We utilized the CHI-TAXI dataset, which was not included in the pre-training phase for any of the three models, for our evaluation. Detailed settings are provided in the Appendix. The results in Table~\ref{tab:compare_large} show that \model\ maintains a significant performance advantage over other advanced large-scale spatio-temporal models. Additionally, \model\ and UniST exhibit notable efficiency improvements compared to UrbanGPT. This may be due to UrbanGPT's reliance on the LLM for making predictions through a question-and-answer format, which hinders its ability to process batch data efficiently. The proposed \model\ model achieves a win-win in performance and efficiency, highlighting its potential as a powerful large-scale model for traffic benchmarks.

\vspace{-0.05in}
\begin{table}[htbp]
\renewcommand\arraystretch{0.90}
    \centering
    % \small
    % \caption{\model's Superior Efficiency and Effectiveness over Large Spatio-Temporal Pre-trained Models.}
    \caption{Comparison of large-scale ST pre-trained models.}
    \vspace{-0.15in}
    \label{tab:compare_large}
    \scalebox{0.95}{
    \begin{tabular}{l|c c c}
        \hline
        \multirow{2}{*}{Model} & \multicolumn{3}{c}{CHI-TAXI}\\
        \cline{2-4}
        & MAE & RMSE & Cost (seconds)\\
        \hline
        UniST & 2.94 & 7.88 & 1.3 \\
        UrbanGPT & 3.26 & 7.10 & $4.5\times 10^4$ \\
        $\textbf{\model}_\text{mini}$ & \textbf{1.74} & \textbf{3.80} & 1.5\\
        \hline
    \end{tabular}
    \vspace{-0.20in}
    }
\end{table}
\vspace{-0.05in}
\vspace{-0.1in}
\section{Related Work}
\label{sec:relate}

\noindent \textbf{Deep Urban Traffic Prediction Models}. 
The rise of deep learning has propelled deep spatio-temporal (ST) models to the forefront of the urban computing, known for their exceptional performance. These models analyze temporal and spatial correlations within historical data to forecast future trends. Prevalent temporal frameworks include recurrent neural networks (RNN)\cite{yu2017deep}, temporal convolutional networks (TCN)\cite{wu2019graph}, and attention networks~\cite{zheng2020gman}. Spatial correlations are typically encoded using graph convolution (GCN)\cite{yu2018spatio}, graph attention (GAT)\cite{zheng2020gman}, and hypergraph neural networks (HGNN)~\cite{yan2023spatio}. Ongoing research refines these temporal and spatial encoding strategies, including multi-scale~\cite{wang2020traffic} and multi-granularity learning~\cite{Guo2019attention}, as well as adaptive~\cite{bai2020adaptive, wu2020connecting} and dynamic graph modeling~\cite{han2021dynamic,zhao2023dynamic}. Researchers also explore cutting-edge techniques, such as neural ordinary differential equation for continuous time modeling~\cite{fang2021spatial, jin2023mtgode} for advancing the spatio-temporal learning. \\\vspace{-0.12in}

\noindent \textbf{Spatio-Temporal Self-Supervised Learning}.
Self-supervised learning (SSL) has emerged as an effective augmentation strategy for spatio-temporal learning. Existing SSL paradigms can be broadly categorized into: (1) Spatio-Temporal Contrastive SSL, which leverages contrastive learning to capture spatial and temporal correlations \cite{li2022mining,zhang2023automated}; (2) Spatio-Temporal Generative SSL, utilizing generative pretext tasks to model the underlying spatio-temporal dynamics via masked autoencoding \cite{shao2022pretraining,li2023gpt}; and (3) Spatio-Temporal Predictive SSL, incorporating auxiliary tasks that predict future spatio-temporal patterns with heterogeneity-aware data augmentation \cite{ji2023spatio}. While these approaches aim to enhance the performance and generalization of spatio-temporal learning models, they are still limited by the constraints of zero-shot forecasting capabilities. \\\vspace{-0.12in}

% Recent advances in pre-training technology have significantly impacted various fields~\cite{devlin2019bert,dosovitskiy2021an, he2022masked}, with ST pre-training emerging as a key area of interest in the ST domain. The primary frameworks for ST pre-training are based on two paradigms: generative-based~\cite{shao2022pretraining,li2023gpt,li2024flashst} and contrastive-based~\cite{liu2022stgcl,zhang2023automated}. Generative pre-training typically involves mask reconstruction or predicting subsequent sequences, while contrastive pre-training focuses on drawing positive examples closer and distancing negative examples to complete the training process. These models are dedicated to generating robust ST representations to achieve performance or efficiency gains in downstream tasks. Despite achieving enormous success, these pre-trained models have not demonstrated zero-shot generalization capabilities for unseen scenarios. This challenging task still requires further research and exploration.

\noindent \textbf{Leveraging Large Language Models in Urban Computing}.
The emergence of large language models (LLMs) has prompted research integrating their benefits into urban computing through three main paradigms: (1) Works like STLLM \cite{liu2024stllm} and TPLLM \cite{ren2024tpllm} employ LLM architectures as ST encoders, fine-tuning for ST representation learning. (2) These efforts leverage LLMs' linguistic capabilities to enhance predictions, using in-context learning for model generalization enhancement \cite{li2024urbangpt} and interpretable generation \cite{guo2024r2tllm}. (3) LLMs as ST Agents: This approach bridges LLMs with tools to facilitate traffic-related queries and reasoning based on human instructions \cite{da2024openti}. While LLMs can improve spatio-temporal model performance, two key limitations exist: \emph{First}, Computational Demands: LLMs' high computational requirements reduce efficiency, posing challenges for real-world deployment. \emph{Second}, Reliance on POI Data: Most approaches heavily rely on the manually-collected, rich textual Point-of-Interest (POI) information, limiting generalization. To address these limitations, there is a need to develop more efficient and highly-generalized models that can learn universal spatio-temporal patterns across diverse transportation scenarios.

\section{Conclusion}
\label{sec:conclusoin}
The work introduces \model, a scalable spatio-temporal foundation model for traffic prediction that achieves precise zero-shot prediction performance across multiple traffic forecasting scenarios. By employing the Transformer encoder architecture as the backbone to model dynamic spatio-temporal dependencies and pre-training on the large-scale traffic datasets, \model\ demonstrates exceptional zero-shot predictive performance on various downstream tasks, matching the results of state-of-the-art baseline models in full-shot scenarios. The proposed \model\ framework can effectively handle data with varying distributions and boasts high computational efficiency. Along with the promising scaling laws observed, this paves the way for the development of a powerful, generalized traffic prediction solution that can be readily applied to diverse urban environments and transportation networks.
% The designed \model\ framework can effectively handle data with varying distributions, along with the promising scaling laws observed, paves the way for the development of a powerful, generalized traffic prediction solution that can be readily applied to diverse urban environments and transportation networks. 
% 

% Moreover, OpenCity possesses high computational efficiency.

\clearpage

\bibliographystyle{ACM-Reference-Format}
\balance
\bibliography{refers} 
% % \bibliography{sample-base}

\clearpage
\appendix \section{Appendix}
\balance
\label{sec:appendix}
% In the appendix section, we offer comprehensive discussion of the experimental setup. This includes detailed dataset information, hyperparameter configuration, experimental setting for the instruction-tuning phase and test phase, as well as the baselines. Furthermore, we present a case study of our \model, showcasing its effectiveness for zero-shot spatio-temporal predictions.

% In the appendix section, we have supplemented the training and optimization process of the \model\ model, detailed data information, settings under different experiments, hyperparameter configurations, baseline explanations, evaluation metrics, and descriptions of deployment experiments.
In the appendix, we have provided comprehensive details on the \model\ model, including the training and optimization procedures, dataset information, experimental settings, hyperparameter configurations, baseline explanations, evaluation metrics, and descriptions of the deployment experiments. This supplementary material aims to offer readers a thorough understanding of the technical aspects and experimental setup behind our work.

\subsection{\bf Model Training and Optimization}
\label{sec:loss_function}
% To learn robust spatio-temporal representations, \model\ leverages multiple large-scale traffic datasets for pre-training. During the training process, we randomly complete one step of training using data from a specific dataset, with one training epoch covering all available training data. \model\ adopts a supervised training paradigm to complete pre-training. After obtaining the output from the L-th layer of the spatio-temporal encoding network, we flatten the features and use a simple linear layer to obtain the prediction $\textbf{Y} \in \mathbb{R}^{R\times T}$. Following previous research~\cite{}, we use the absolute error as the loss function $\mathcal{L}$, formalized as follows:
To learn robust spatio-temporal representations, our model leverages multiple large-scale traffic datasets for pre-training. This approach allows the model to capture the inherent complexities and dependencies within transportation networks by leveraging diverse data sources. During the training process, we randomly complete one step of training using data from a specific dataset, with one training epoch covering all available training data. Our model adopts a supervised training paradigm to complete this pre-training stage.

After obtaining the output from the $L$-th layer of the spatio-temporal encoding network, we flatten the features and use a linear layer to obtain the prediction $\textbf{Y} \in \mathbb{R}^{R\times T}$ Following previous research~\cite{bai2020adaptive,li2024flashst}, we use the absolute error as the loss function $\mathcal{L}$:
\begin{align}
    \label{eq:loss}
    \mathcal{L} = \frac{1}{RT}\sum_{r=1}^R\sum_{t=1}^T |(\hat{\textbf{Y}}_{r,t} - \textbf{Y}_{r,t})|
\end{align}
Our model is designed to capture the intricate spatial and temporal dependencies inherent in transportation networks by leveraging multi-dataset pre-training. This approach enables the model to learn robust spatio-temporal representations that can generalize well to a wide range of scenarios, including spatial generalization with zero-shot prediction and temporal generalization with long-term forecasting. By pre-training on diverse datasets, the model can extract meaningful features and patterns that are transferable across different spatial and temporal domains, enhancing its versatility and performance in real-world traffic forecasting tasks.

\subsection{\bf Comprehensive Experimental Settings}
\subsubsection{\bf Data Sources and Characteristics}
To comprehensively evaluate the generalization capabilities and predictive performance of the \model\ approach, we conducted training and testing across a diverse set of large-scale public real-world datasets. These datasets cover multiple traffic-related data categories from various regions, including the United States and China, spanning several major cities such as New York City, Chicago, Shanghai, and others. \textbf{(i) Traffic flow Data}~\cite{liu2023largest,Song2020spatial}: CAD-X and PEMS-X from California; \textbf{(ii) Taxi Demand Data}~\cite{li2024urbangpt}: X-TAXI from New York City and Chicago; \textbf{(iii) Bicycle Trajectory Data}~\cite{li2024urbangpt}: X-BIKE from New York City; \textbf{(iv) Traffic Speed Statistics}~\cite{yuan2024unist}: Traffic-X from major cities in China; \textbf{(v) Traffic Speed Statistics}~\cite{li2017diffusion,yu2018spatio}: METR-LA, PEMS-BAY, and PEMS07M from Los Angeles, the Bay Area, and California; \textbf{(vi) Traffic Index Statistics}~\cite{lu2022spatio}: X-DIDI from Shenzhen and Chengdu.

The taxi demand datasets, bicycle trajectory datasets, and traffic speed statistics datasets (Traffic-X) used in this study are grid-based in nature, while the remaining datasets are sensor-based. To distinguish between the specific districts or cities represented within each dataset, the ``X'' placeholder is utilized. 

The detailed statistics for all datasets used in this paper is shown in Table~\ref{tab:dataset}. Considering that a large number of regions might lead to excessive GPUs memory overhead, we segmented datasets CAD4, CAD7, CAD8, and CAD12 according to the order of nodes. Specifically, the number of regions in CAD4-1, CAD4-2, CAD4-3, and CAD4-4 are 621, 610, 593, and 528; in CAD7-1, CAD7-2, and CAD7-3 are 666, 634, and 559; in CAD8-1 and CAD8-2 are 510 and 512; and in CAD12-1 and CAD12-2 are 453 and 500. Additionally, the last column of Table~\ref{tab:dataset} indicates whether each dataset was utilized for pre-training in \model. To evaluate the model's performance in a supervised setting, we reserved a portion of the data from the datasets involved in the supervised experiments for evaluation.

\begin{table*}[t]
\renewcommand\arraystretch{1.6}
    \centering
    \small
    \caption{Statistical information of the datasets.}
    \vspace{-0.05in}
    \label{tab:dataset}
    \scalebox{0.995}{
    \begin{tabular}{c c c c c c c c c}
        \hline
        \multicolumn{1}{c}{\textbf{Dataset}} & \textbf{Category} & \textbf{Geoinformation} & \textbf{\# Regions} & \textbf{Sample rate} & \textbf{Time span(Y/M/D)} & \textbf{Obs.} & \textbf{Traffic net} & \textbf{For pre-training}\\
        % \cline{2-17}
        \hline
        \multicolumn{1}{c}{CAD3} & Traffic flow & California, USA & 480 & 5 minutes & 2020/03/01-2020/04/30 & 8.4M & Sensor & No  \\
        \multicolumn{1}{c}{CAD4} & Traffic flow & Bay Area, USA & 2352 & 5 minutes & 2020/01/01-2020/03/15 & 50.8M & Sensor & Yes  \\
        \multicolumn{1}{c}{CAD5} & Traffic flow & California, USA & 211 & 5 minutes & 2020/03/01-2020/04/30 & 3.7M & Sensor & No  \\
        \multicolumn{1}{c}{CAD7} & Traffic flow & Los Angeles, USA & 1859 & 5 minutes & 2020/01/01-2020/03/15 & 40.2M & Sensor & Yes  \\
        \multicolumn{1}{c}{CAD8} & Traffic flow & Los Angeles, USA & 1022 & 5 minutes & 2020/01/01-2020/03/15 & 22.1M & Sensor & Yes  \\
        \multicolumn{1}{c}{CAD12} & Traffic flow & Los Angeles, USA & 953 & 5 minutes & 2020/01/01-2020/03/15 & 20.6M & Sensor & Yes  \\
        \multicolumn{1}{c}{PEMS04} & Traffic flow & California, USA & 307 & 5 minutes & 2018/01/01-2020/02/28 & 5.2M & Sensor & Yes  \\
        \multicolumn{1}{c}{PEMS08} & Traffic flow & California, USA & 170 & 5 minutes & 2016/07/01-2020/08/31 & 3.0M & Sensor & Yes  \\
        \multicolumn{1}{c}{NYC-TAXI} & Taxi demand & New York City, USA & 263 & 30 minutes & 2016/01/01-2021/12/31 & 27.7M & Grid & Yes  \\
        \multicolumn{1}{c}{CHI-TAXI} & Taxi demand & Chicago, USA & 77 & 30 minutes & 2021/01/01-2021/12/31 & 1.3M & Grid & No \\
        \multicolumn{1}{c}{NYC-BIKE} & Bike trajectory & New York City, USA & 540 & 30 minutes & 2020/01/01-2020/12/31 & 9.5M & Grid & No  \\
        \multicolumn{1}{c}{TrafficZZ} & Traffic speed & Zhengzhou, China & 676 & 30 minutes & 2022/03/05-2022/04/05 & 0.9M & Grid & Yes  \\
        \multicolumn{1}{c}{TrafficHZ} & Traffic speed & Hangzhou, China & 672 & 30 minutes & 2022/03/05-2022/04/05 & 0.9M & Grid & Yes  \\
        \multicolumn{1}{c}{TrafficCD} & Traffic speed & Chengdu, China & 728 & 30 minutes & 2022/03/05-2022/04/05 & 1.0M & Grid & Yes  \\
        \multicolumn{1}{c}{TrafficJN} & Traffic speed & Jinan, China & 576 & 30 minutes & 2022/03/05-2022/04/05 & 0.8M & Grid & Yes  \\
        \multicolumn{1}{c}{TrafficSH} & Traffic speed & Shanghai, China & 896 & 30 minutes & 2022/03/05-2022/04/05 & 1.3M & Grid & No  \\
        \multicolumn{1}{c}{METR-LA} & Traffic speed & Los Angeles, USA & 207 & 5 minutes & 2012/03/01-2022/04/30 & 3.6M & Sensor & Yes  \\
        \multicolumn{1}{c}{PEMS-BAY} & Traffic speed & Bay Area, USA & 325 & 5 minutes & 2017/01/01-2017/04/30 & 11.2M & Sensor & Yes  \\
        \multicolumn{1}{c}{PEMS07M} & Traffic speed & California, USA & 228 & 5 minutes & 2017/05/01-2017/08/31 & 2.9M & Sensor & No  \\
        \multicolumn{1}{c}{SZ-DIDI} & Traffic index & Shenzhen, China & 627 & 10 minutes & 2018/01/01-2018/02/28 & 5.3M & Sensor & No  \\
        \multicolumn{1}{c}{CD-DIDI} & Traffic index & Chengdu, China & 524 & 10 minutes & 2018/01/01-2018/02/28 & 4.5M & Sensor & No  \\
        \hline      
        % \hline
    \end{tabular}
    }
    % \vspace{-0.1in}
\end{table*}

\subsubsection{\bf Hyperparameter Configuration}
We release three different versions of the \model\ model, each with varying parameter counts: $\textbf{\model}_\text{mini}$ (\textbf{2M} parameters), $\textbf{\model}_\text{base}$ (\textbf{5M} parameters), and $\textbf{\model}_\text{plus}$ (\textbf{26M} parameters). We scale the \model\ architecture by increasing the dimensions of the hidden layers and the number of layers in the spatio-temporal encoder, as detailed in Table 1. For our long-term traffic forecasting focus, we set the historical and future time spans ($H$ and $F$) to 1 day. This configuration is determined by the aggregation frequency - when the frequency is 5 minutes, $H=F=288$. Furthermore, we set the patch length $P$ and stride $S$ to one-hour time spans, which correspond to a value of 12 when the aggregation frequency is 5 minutes.

The regional embedding dimensionality $k$ is set to 8 and the balancing weight $\alpha$ is set to 0.05. The dropout ratios for the attention matrix $\delta_a$ and the spatio-temporal network $\delta$ are set to 0.3 and 0.1, respectively. We maximize the batch size settings based on the GPU memory usage across the different models. Additionally, the hyperparameters for all baseline configurations adhere to the settings provided in the original papers or the official released codes. All training and testing are conducted on a server equipped with 8 $\times$ NVIDIA A100-SXM4-40GB GPUs. 

% Further details on the hyperparameter settings are provided in the supplementary materials.

\subsubsection{\bf Detailed Experimental Setup.} The detailed settings of different experiments and the division of datasets are as follow:

\emph{Zero-shot Evaluation Setup}: In the zero-shot evaluation, the proposed \model\ model is directly used for testing, while the baselines are trained and tested in a supervised setting. For the traffic flow datasets (CAD3, CAD5) and the traffic speed datasets (PEMS07M, TrafficSH), the split ratios for training, validation, and testing sets are 0.5, 0.1, and 0.4, respectively. For the taxi demand dataset (CHI-TAXI) and the bicycle trajectory dataset (NYC-BIKE), these ratios are 0.2, 0.2, and 0.6. All the above datasets are unavailable during the pre-training phase of \model.

\emph{Supervised Evaluation Settings}: In the supervised scenario, the pre-trained \model\ model is directly employed for evaluation. Contrary to the zero-shot setting, all these datasets are included in the pre-training phase. For the CAD8-1, CAD8-2, and CAD12-2 datasets, the proportions for training, validation, and testing are set at 0.8, 0.1, and 0.1, respectively. For the PEMS-BAY dataset, these ratios are 0.5, 0.1, and 0.4. In the case of the NYC-TAXI dataset, all data from 2016 to 2020 are used as the training set, data from the first two months of 2021 serves as the validation set, and data from the remaining ten months of 2021 is used as the test set.

\emph{Fast adaptability Evaluation Setup}: In this experiment, we tested the \model\ model using data categories not included in its training data, specifically the traffic index data from CD-DIDI and SZ-DIDI. The division ratios for training, validation, and testing were set at 0.5, 0.1, and 0.4, respectively. We utilize the 'Efficient Fine-tuning' approach, in which only the prediction head of the model (the last linear layer) is updated for a maximum of 3 training epochs. For baseline comparisons, we set the maximum number of training epochs at 100 and use an early stopping criterion after 15 epochs, without freezing any parameters. To maintain consistent timing expense statistics, we set the batch size at 64 for all models (except for PDFormer due to its higher GPU memory overhead).

In addition, we utilize $\text{\model}_\text{plus}$ as the backbone model for our \emph{ablation experiments} to assess the effectiveness of various modules. For \emph{comparisons with large-scale spatio-temporal pre-trained models}, we directly use the open-source large model from the source repository to conduct performance evaluations. Taking into account the configurations of UrbanGPT and UniST, we employ the last month of CHI-TAXI data as the test set and evaluate the prediction accuracy of the first six steps across different models.

\subsubsection{\bf Evaluation Metrics for Traffic Prediction}
The experiments utilize three widely used evaluation metrics in traffic forecasting and regression tasks: Mean Absolute Error (MAE), Root Mean Square Error (RMSE), and Mean Absolute Percentage Error (MAPE). MAE measures the average absolute difference between the predicted values and true labels, RMSE captures the standard deviation of the prediction errors, and MAPE represents the average absolute percentage difference between the predicted values and true labels. Lower values for these metrics indicate better performance of the model, as they quantify the errors between the predicted values and true labels from different perspectives.

\subsubsection{\bf Baseline Description}
We selected 10 advanced spatio-temporal prediction models as baselines, all of which have demonstrated considerable success in traffic prediction tasks. Given that most of these models are primarily designed for short-term predictions, we have enhanced their long-term prediction capabilities by applying patch embedding, as introduced in Section~\ref{sec:patch_eb}. The selected baseline models fall into the following three categories:
\noindent \textbf{RNN-Based Spatio-temporal Prediction Models:}
\begin{itemize}[leftmargin=*]
\item \textbf{AGCRN}~\cite{bai2020adaptive}: This model incorporates learnable node embeddings into recurrent neural networks (RNNs), enabling it to capture the region-specific spatio-temporal evolution patterns.
\item \textbf{MSDR}~\cite{liu2022msdr}: This method enhances RNNs by addressing the issue of long-term information forgetting. It achieves this by preserving multi-step hidden states at each time unit, facilitating the modeling of both long and short-term temporal relationships.
\end{itemize}

\noindent \textbf{Attention-Based Spatio-temporal Neural Networks:}
\begin{itemize}[leftmargin=*]
\item \textbf{ASTGCN}~\cite{Guo2019attention}: This model leverages an attention mechanism to develop multi-granular spatiotemporal representations.
\item \textbf{STWA}~\cite{cirstea2022towards}: In this method, time-varying and location-specific parameters are integrated into the attention mechanism to capture time-aware and region-aware spatio-temporal dependencies.
\item \textbf{PDFormer}~\cite{jiang2023pdformer}: This method employs attention networks to learn ST representations separately in the temporal and spatial dimensions. Delay-aware modeling is incorporated into the spatial attention mechanism to model traffic mobility patterns.
\end{itemize}

\noindent \textbf{RNN-Based Spatiotemporal Prediction Models:}
\begin{itemize}[leftmargin=*]
\item \textbf{TGCN}~\cite{Zhao2020TGCN}: This model employs GNNs to capture spatial correlations and utilizes RNNs to model temporal dependencies.
\item \textbf{STGCN}~\cite{yu2018spatio}: In this method, graph convolutional networks and gated temporal networks are utilized to model spatial dependencies and temporal dependencies, respectively.
\item \textbf{GWN}~\cite{wu2019graph}: This model uses a learnable graph structure and 1-D dilated convolution to capture spatial and temporal correlations.
\item \textbf{MTGNN}~\cite{wu2020connecting}: This method employs a learnable graph neural network to model spatial dependencies, and the diffusion convolution network is used to learn multi-scale temporal correlations.
\item \textbf{STSGCN}~\cite{Song2020spatial}: It introduces a spatio-temporal graph construction method to simultaneously learn temporal and spatial correlations.
\end{itemize}

\subsection{\bf Model Deployment}
In this section, we examine the deployment-related challenges associated with \model. Given that the datasets for all our experiments are derived from real-world scenarios and align with actual prediction environments, the performance of our model in deployment settings has been validated in the aforementioned experiments. Our focus here extends to assessing the efficiency of \model\ in practical applications. In contrast to offline prediction, online evaluation typically involves processing a single batch of predictions within a specific timeframe, dictated by the regular intervals at which traffic data is updated and aggregated (\eg, every 5 minutes, every 2 hours, 1 day). We deployed \model\ on a single NVIDIA A100-SXM4-40GB GPU and evaluated its prediction speed across cities of varying sizes. The task involved predicting the next day's traffic trends based on the data from the previous day.

According to the results presented in Table~\ref{tab:deployment}, \model\ is capable of delivering predictions in an exceptionally short period. Even in cities with numerous regions and fine-grained prediction requirements, \model\ completes individual predictions in less than 3 seconds. This remarkable efficiency not only demonstrates the model's robustness but also reinforces \model's potential to serve as a foundational model for traffic prediction.

\begin{table}[t]
\renewcommand\arraystretch{1.1}
    \centering
    % \small
    \caption{Deployment study of \model\ model.}
    \vspace{-0.05in}
    \label{tab:deployment}
    \scalebox{1}{
    \begin{tabular}{l|c c c c c}
        \hline
        \multirow{2}{*}{Model} & \multicolumn{5}{c}{$\textbf{\model}_\text{plus}$}\\
        \cline{2-6}
        & MAE & RMSE & Cost (s) & Obs. & \# Regions \\
        \hline
        CHI-TAXI & 1.67 & 3.53 & 2.00 & 3,936 & 77\\
        NYC-BIKE & 3.88 & 6.29 & 2.10 & 25,920 & 540\\
        CAD3 & 30.68 & 48.50 & 2.08 & 138,240 & 480\\
        TrafficSH & 0.51 & 0.78 & 2.22 & 43,008 & 896\\
        \hline
    \end{tabular}
    \vspace{-0.1in}
    }
\end{table}

\end{document}